%% file: xtda.tex
\documentclass[journal]{IEEEtran}

\usepackage{booktabs}
\usepackage{multirow}
\usepackage{textcomp}
\usepackage{algorithmic}
\usepackage{algorithm}
\usepackage{array}
\usepackage[caption=false,font=footnotesize]{subfig}

\usepackage{stfloats}
\usepackage{url}
\usepackage{verbatim}

\usepackage{amsfonts}
\usepackage{amssymb,amsmath,amsthm}
\usepackage{mathtools}
\newtheorem{definition}{Definition}

\theoremstyle{definition}

\usepackage{graphicx}
\usepackage{duckuments}
\usepackage{wrapfig}

\usepackage{csquotes}
\usepackage{hyperref}

\usepackage{tikz}
\usetikzlibrary{calc,intersections,through,backgrounds}

\usepackage{cite}
\usepackage[algo2e]{algorithm2e}

\usepackage{bbm}

\usepackage{balance}

\newcommand{\G}{\mathcal{G}}
\newcommand{\V}{\mathcal{V}}

\begin{document}
 
\title{Explaining the Power of Topological Data Analysis in Graph Machine Learning}

\author{Funmilola Mary Taiwo, Umar Islambekov, Cuneyt~Gurcan~Akcora
\thanks{F.M Taiwo is with the Department of Statistics, University of Manitoba}
\thanks{U. Islambekov is with the Department of Mathematics, Bowling Green State Uni., Bowling Green, OH, USA}
\thanks{C.G Akcora is with AI Institute, University of Central Florida}
}

\IEEEoverridecommandlockouts
\IEEEpubid{\makebox[\columnwidth]{0000--0000/00\$00.00~\copyright2024 IEEE \hfill} \hspace{\columnsep}\makebox[\columnwidth]{ }}

\maketitle

\IEEEpubidadjcol

\begin{abstract}

Topological Data Analysis (TDA) has been praised by researchers for its ability to capture intricate shapes and structures within data. TDA is considered robust in handling noisy and high-dimensional datasets, and its interpretability is believed to promote an intuitive understanding of model behavior. However, claims regarding the power and usefulness of TDA have only been partially tested in application domains where TDA-based models are compared to other graph machine learning approaches, such as graph neural networks.

We meticulously test claims on TDA through a comprehensive set of experiments and validate their merits. Our results affirm TDA's robustness against outliers and its interpretability, aligning with proponents' arguments. However, we find that TDA does not significantly enhance the predictive power of existing methods in our specific experiments, while incurring significant computational costs. We investigate phenomena related to graph characteristics, such as small diameters and high clustering coefficients, to mitigate the computational expenses of TDA computations. Our results offer valuable perspectives on integrating TDA into graph machine learning tasks.

\end{abstract}

\begin{IEEEkeywords}
Topological Data Analysis, Persistent Homology, Mapper
\end{IEEEkeywords}

\section{Introduction}
\IEEEPARstart{T}{opological} Data Analysis is gaining much traction in the fields of statistics, computer science, and mathematics. The underlying concept of TDA is understanding the shape of data which focuses on detecting and encoding topological features such as connected components, loops, voids, etc., that are present in datasets in order to improve inference and prediction~\cite{CarlssonTopologyData2009}.
\input{introduction}

\section{Problem Formulation}\label{sec:background}
\noindent\input{background}

\subsection{Research Questions}\label{sec:researchq} 
We attempt to answer the following questions:
\begin{enumerate}
\itemsep0em 
\item \textbf{Power:} What are the TDA-based methods' performance limits in graph classification? How do these limits compare to those of graph kernels and graph neural networks?
\item \textbf{Components:} What specific elements of TDA demonstrate predictive capabilities in graph classification? What types of complexity, vectorization methods, and topological features are relevant in this context?
\item \textbf{Complementary Power:} Can TDA-based methods be combined with traditional graph features to improve performance? How can we effectively integrate TDA-based methods with other methods to capture both topological and geometric features of graphs?
\item \textbf{Scalability:} How do TDA-based methods scale with increasing graph size and complexity? Can these methods handle large graphs with millions or billions of nodes and edges?
\item \textbf{Robustness:} How robust are TDA-based methods to noise and perturbations in the input data?  
\item \textbf{Interpretability:} How interpretable are TDA-based methods compared to other graph machine learning methods such as graph neural networks? Can we gain insights into the underlying structure of the data and the features that are most important for prediction using TDA-based methods?
\end{enumerate}

\section{Persistence-based Topological Methods}\label{sec:tutorial}
\noindent\input{topology}

\section{Experimental Setup} \label{sec:experiments}
\noindent This section explains the experimental setup to answer our research questions. Complex types, filtrations and vectorization schemes --- all essential components of TDA --- are discussed in Section~\ref{sec:RQanswers} when answering the questions.   We analyze the performance of TDA and kernel methods on nine benchmark graph classification datasets~\cite{KKMMN2016}. 

The datasets $\texttt{BZR}$, $\texttt{COX2}$, $\texttt{DHFR}$, $\texttt{NCI1}$, $\texttt{PROTEINS}$, $\texttt{MUTAG}$, $\texttt{ENZYMES}$ are graphs obtained from medical or biological frameworks while $\texttt{REDDIT-MULTI-5K}$ (RED-5K) and $\texttt{REDDIT-MULTI-12K}$ (RED-12K) are derived from social networks. Appendix~\ref{AppendixA} contains the statistics of these datasets. 

The baseline graph features are graph density, graph diameter, clustering coefficient, spectral gap, node assortativity, number of cliques in the graph, number of disconnected components, and three-vertex motif counts. See Appendix~\ref{sec:graphFeatures} for details.

We use a Random Forest classifier because it is inherently explainable~\cite{du2019techniques}, i.e.,  we can analyze tree splits to learn which features are useful in prediction. This, in turn, allows us to pinpoint which features are informative. 

\noindent\textbf{Implementation: } We have implemented our methods in Python and shared them on GitHub.\footnote{https://github.com/Phunmey/XTDA-paper} We construct VR and Alpha complexes on distance matrices by using  {\fontfamily{qcr}\selectfont Ripser}~\cite{Bauer2021Ripser} and {\fontfamily{qcr}\selectfont GUDHI}~\cite{gudhi:AlphaComplex} respectively, which are persistent homology libraries in Python available under MIT license. We have performed all our analyses on a Digital Research Alliance of Canada (\url{https://ccdb.computecanada.ca}) server which hosts 2 x AMD Rome 7532 @ 2.40 GHz 256M cache L3 and 8000MB memory per CPU.  

In each experiment, we divide the data into 80\% for training and 20\% for testing. We train and evaluate models using the Random Forest algorithm from the Scikit-learn library. The Random Forest model is trained using GridSearchCV to tune hyperparameters, specifically the number of estimators ($n\_estimators$) in the range of 200 to 500, maximum depth ($max\_depth$) in the range of 2 to 10, and with a cross-validation value of 10. To ensure robustness, we repeat the experiments 10 times and report both the average values and standard deviations of the results.

\noindent\textbf{Metrics:} The metrics we use in evaluating our models are accuracy, RMSE, MAE and R2\_Score (see Appendix~\ref{sec:metrics} for details).
\vspace{-10px}

\section{Answers to Research Questions}\label{sec:RQanswers}
\noindent Before individually examining each of the six research questions, we will address the power and component parts together. The reason for this approach is that these two aspects serve as the fundamental basis for addressing the remaining questions.

\subsection{Power and Components of TDA}

The application of the VR complex is straightforward, where we use the shortest-path length or resistance distance (as described in Section~\ref{sec:background}) as the distance metric between pairs of vertices. While there are various distance functions available (such as weighted distance) for graph vertices, most of our graph datasets, such as $\texttt{PROTEINS}$, are unweighted and undirected, making such distance metrics unsuitable. For both the VR and Alpha complex algorithms, we compute the persistence homology in dimensions $p=0$ and $p=1$. 

To construct an AC filtration, we need to embed our metric space data into a Euclidean space $\mathbb{R}^m$, as direct construction is not feasible. This involves finding a representation of $k$ points in $\mathbb{R}^m$ (in our case, $m=2$) that preserves the pairwise distances between the points in the original metric space. To this end, we can utilize dimensionality reduction methods such as t-distributed Stochastic Neighborhood Embedding (t-SNE), Multidimensional Scaling (MDS), and Principal Component Analysis (PCA). The high computational complexity of t-SNE which is around $O(|V|^2)$~\cite{pezzotti2016approximated}, makes it unsuitable for large graphs. Thus, we were constrained to use MDS and PCA in our experiments, of which we present the results obtained using PCA since it is the fastest method.  

In order to meet the requirement of dimensionality reduction methods for finite distance values, we apply PCA to each connected component of disconnected graphs and merge the resulting matrices. We construct a filtration to generate a data structure known as a simplex tree, which represents the filtered simplicial complex. Subsequently, we vectorize the resulting persistence diagrams accordingly. The resulting feature matrix is then utilized as input for our graph classification task.

We start by noting a crucial point for TDA on graphs. Given a non-null graph, traditional approaches query the graph for a descriptor (e.g., average clustering coefficient or a histogram of vertex colouring), always yielding a result. However, in the case of TDA, it's possible that no meaningful results are produced (i.e., no holes of dimension 1 exist), indicating that TDA might not identify any significant structures to quantify. In other words, power can only exist when the topology allows for it. For that reason, we have studied but do not report all results on vectorizations of $\geq 2$-dim holes as such holes rarely existed.

We analyze TDA components along three fundamental dimensions: filtration, complex type,  and vectorization type. Filtration plays a crucial role in extracting useful information via Persistent Homology. It encompasses three key aspects: range, length, and step size. Regarding the filtration range, we set the start value ($\epsilon_1$) to 0 and the end value ($\epsilon_n$) to the largest finite value (i.e., 1) in the (normalized) distance matrix. To ensure that all topological features are captured, we divide the range into smaller intervals using a predefined list of random numbers, denoted as $n \in \{10, 20, 50, 100\}$. These values serve as filtration step sizes, resulting in a set of filtration thresholds denoted as $\epsilon_i$ for $i = 1, \ldots, n$.  After experimenting with various filtration steps, we have found $n=100$ to provide a fine grained view on the filtration while providing the most accurate models. To illustrate, figure~\ref{fig:bettidistribution} presents a sample plot depicting the distribution of Betti numbers when $n=100$.   The step-size in the filtration process leads us to consider the third aspect, which is the \textit{filtration length}. The filtration length refers to the extent of the filtration thresholds, and it differs from the range and step size in a significant way. The length specifically focuses on the percentage of the range that will be analyzed. The reason for using smaller length values, rather than the full range, is rooted in the fact that PH may not contain any meaningful information beyond a certain point in the filtration range.

Consider figure~\ref{fig:bettidistribution}, where the 1-dimensional and 2-dimensional features cease to exist after reaching the threshold $\epsilon_{60}$. However, as we increase the filtration threshold $\epsilon$ from $\epsilon_{60}$ to $\epsilon_{100}$, the number of simplices increases due to newly added edges, leading to an increased complexity in the PH representation. Nevertheless, this increased complexity does not provide any new insights into 1-dimensional and 2-dimensional holes. Hence, it becomes crucial to determine the filtration length that captures useful information in PH computations.

\noindent\textbf{Components: Complex Type.} Sections~\ref{sec:complex} and ~\ref{sec:apxalphacomplex} presents the two complexes that we consider in a persistent homology~\cite{zomorodian2005zomorodian} setting: the Vietoris-Rips (VR) complex~\cite{vietoris1927hoheren} and the Alpha complex (AC)~\cite{akkiraju1995alpha} in sublevel filtration defined over i) shortest path distance and ii) graph resistance distance.

\noindent\textbf{Components: Filtration.} We set the filtration range from 0 to 1, divided into intervals using predefined step sizes ($n = 10, 20, 50, 100$), generating filtration thresholds $\epsilon_i$ ($i = 1, \ldots, n$). 
 
\noindent\textbf{Components: Vectorization.}  We vectorize the encoded information about the $0-$ and $1-$ dimensional holes using i) Betti functions, ii) persistence landscape and iii) persistence silhouette. 

\noindent\textbf{Power: Accuracy.} Table~\ref{tab:firstexperiment} presents the outcomes of the filtration process, as well as the results obtained using the Weisfeiler-Leman (WL) kernel and the baseline model, utilizing graph features. Among the nine datasets examined, the WL kernel demonstrates the highest accuracy in six of them. On the other hand, the feature baseline model achieves the second highest accuracy in five and the highest accuracy in two datasets. When considering the second highest accurate results, the Vietoris-Rips filtration with Betti functions (VR-B) performs well in two datasets, while the Alpha complex produces the second best results in one dataset. Importantly, our findings indicate that Betti function-based results exhibit considerably higher accuracy values (avg $67.7\%$) compared to silhouette (avg $61.9\%$)  and landscape-based (avg $60.0\%$) results. 

\input{tables/accuracytable.tex}

\noindent\textbf{Power: Stability.} From Table~\ref{tab:firstexperiment}, it is evident that TDA-based classifiers exhibit a significantly higher standard deviation in accuracy values ($3.08$) compared to kernel methods ($0.99$) and the baseline approach ($2.77$). 

\noindent\textbf{Why is the deviation so high?} TDA computations are deterministic, hence the deviations of the classifier must be due to the stochastic nature of the classifier's feature and data point selection during training. However, the deviations are still intrinsically due to TDA. The success of the classifier appears to rely on chance when it comes to picking useful features, such as when a hole in the data disappears at a certain point. However, the main premise of TDA was that persistent holes (i.e., those that stay alive over a longer filtration threshold) are more informative. In that case, persistent holes, which  by definition appear in multiple thresholds, must have higher chances of being picked by the classifier as a feature to be used. Hence, the classifier,which is likely to select informative features, should exhibit stable performance. However, our results, based on the datasets examined in this study, indicate that short-lived features actually convey more useful information.

\subsection{Complementary Power of TDA}

\input{tables/detail01table.tex}
Table~\ref{tab:Betti0and1experiment} provides results on the complementary power of TDA models. Specifically, we use the baseline model which contains traditional graph features and add TDA features to the model. Addition of topological information to the baseline model improves the accuracy values in $\texttt{BZR}$, $\texttt{DHFR}$, $\texttt{ENZYMES}$, $\texttt{MUTAG}$, and $\texttt{PROTEINS}$ datasets. The results demonstrate that 0-dimensional model utilizing data from Baseline and Alpha complex (Base + AC-B) achieves better accuracy than the baseline model on four datasets whereas, Base + AC-B achieves highest accuracy on three datasets in the 1-dimensional models. Furthermore, in three datasets where TDA+baseline has better accuracy than the baseline, the TDA result also has lower standard deviation in accuracy. Our analysis shows that \textbf{TDA models in fact bring complementary power to a baseline model that uses traditional graph features, however the increase in accuracy is trivial}. The improvement is 0.2\% of the accuracy of the baseline model per dataset.

\subsection{Scalability of TDA}

A significant limitation that has hindered the extensive usage of TDA is its high  computational cost (see Section~\ref{sec:complexity}). Consequently, TDA models can be impractical for real-world applications that demand rapid and scalable analysis. Analyzing these computational challenges is crucial to unlocking the full potential of TDA in various domains. To this end, we designed a set of experiments to study these computational limitations and explore strategies to mitigate them in Section~\ref{sec:surrogatemodel}. 

Table~\ref{time: filtration} presents the run time required to compute the persistence diagrams for filtration methods and their vectorizations, fit the Weisfeiler-Leman kernel to our data, and compute all the graph features for the baseline method. 

As indicated in Table~\ref{time: filtration}, both the WL kernel and the baseline method exhibit good scalability on small datasets. Among the filtration methods, Alpha complex-based models (AC-B, AC-L, AC-S) demonstrate favorable scalability compared to Vietoris-Rips (VR). This outcome is expected due to the significant difference in matrix dimension involved in the instance computation of both filtrations. 

\input{tables/timetable.tex}

In terms of scalability, we observe that both graph characteristics-based and Weisfeiler-Leman~ models perform better than TDA-based models. However, the computational costs of Alpha complex are not as high as those of the baseline in large graphs.

\subsection{Robustness of TDA}

The vulnerability of classic machine learning models to data perturbations, known as adversarial examples, is well-known~\cite{zugner2018adversarial}. Even subtle changes to the input can lead to incorrect predictions, raising concerns about the trustworthiness of machine learning predictions. The study of adversarial robustness in graphs has gained attention recently, with the initial focus on node-level classification, demonstrating the susceptibility of graph neural networks (GNNs) to adversarial perturbations~\cite{zugner2018adversarial}. Since then, the field has rapidly expanded, with researchers exploring diverse tasks, models, and strategies to enhance the robustness of GNNs~\cite{gosch2023revisiting}.

In the context of TDA-based models, robustness is often mentioned~\cite{skaf2022topological}; however, to our knowledge, we are the first to rigorously test this assumption. Testing robustness can involve various methods, such as assessing the impact of node/edge additions, deletions, or node feature updates. For our experiment, we focused on edge deletion due to the prevalence of incomplete data scenarios.

\begin{figure*}[!t]
\centering{
\subfloat[Baseline - graph features]{\includegraphics[width=2.25in]{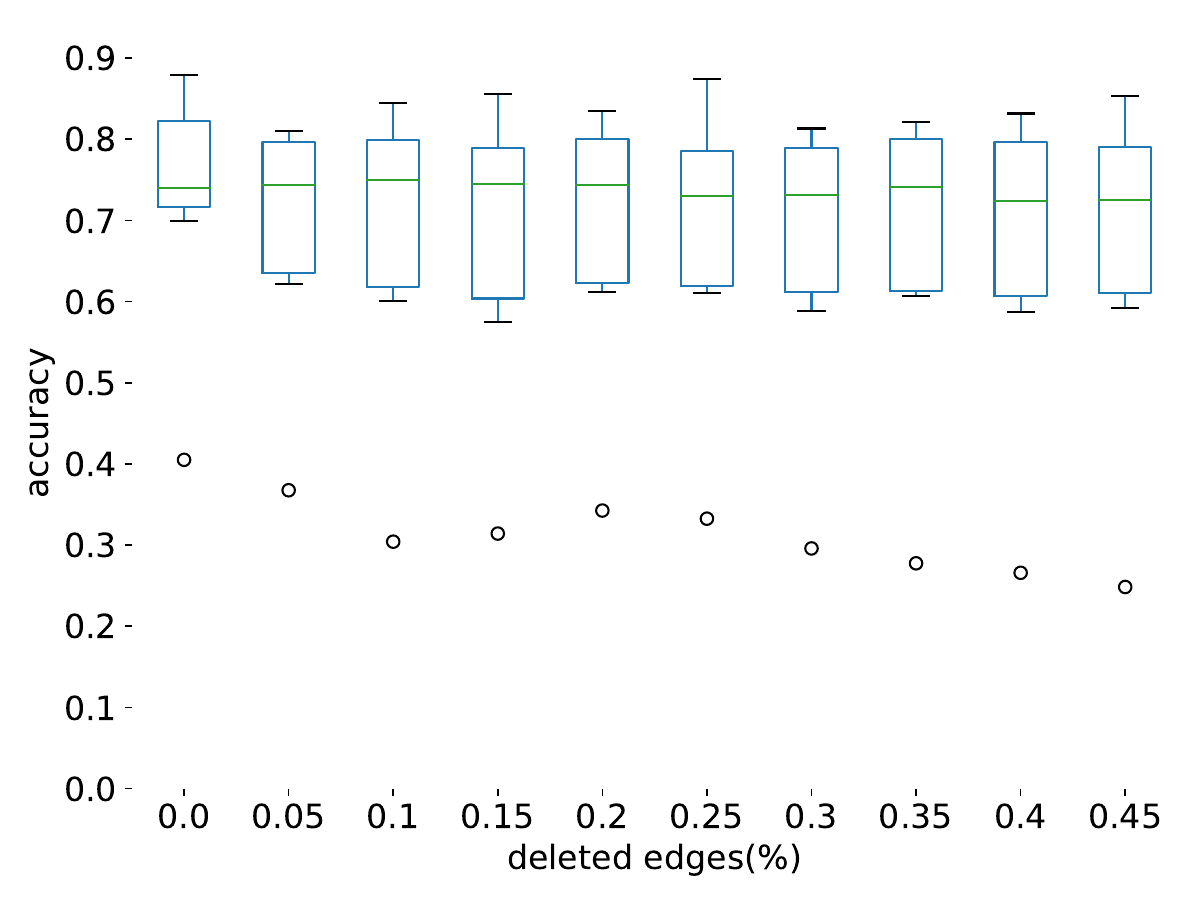}%
     \label{fig:gsbox}}
\subfloat[Weisfeiler-Leman Kernel]{\includegraphics[width=2.25in]{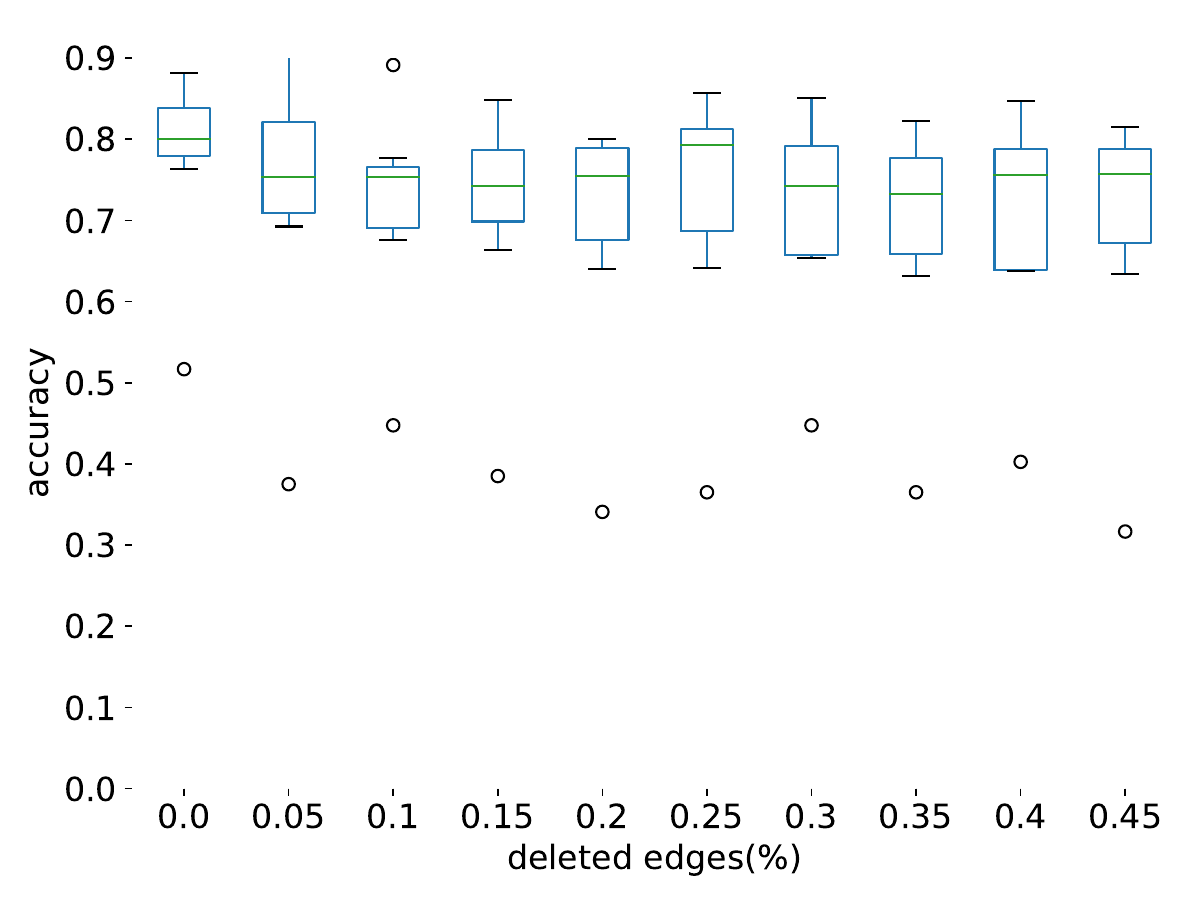}%
     \label{fig:kernelbox}}
\subfloat[VR-Betti with SPD]{\includegraphics[width=2.25in]{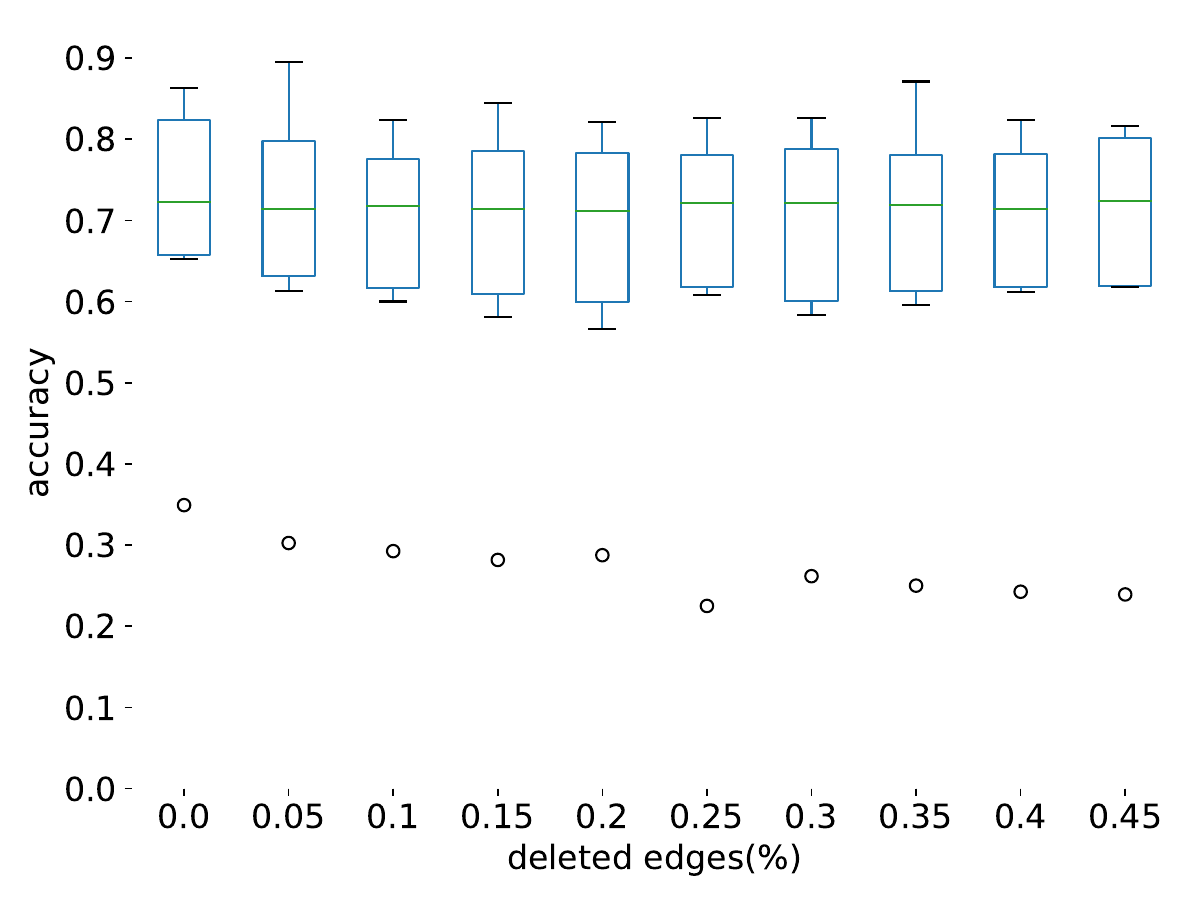}%
     \label{fig:spdbox}}
}
\caption{Variation in Accuracy as a Function of the Percentage of Deleted Edges in a Graph, Averaged Over the Nine Datasets. TDA-Based Vietoris-Rips (VR) Model in Panel C Exhibits Better Decay Rates. For All Three Models, \texttt{ENZYMES} is the Outlier Dataset.}
\label{fig:robustness}
\end{figure*}

We attacked our datasets by randomly removing edges in the range of 0\% to 45\% with an increment of 5\% and report the accuracy results for the methods using boxplots in figure~\ref{fig:robustness}. The figure showcases the variation in accuracy as a function of the percentage of deleted edges in a graph, indicating that TDA-based models, which uses both 0- and 1-dimensional Betti functions, exhibit better decay rates.

Comparing the result of graphs with edge deletion to those of the full graph, figures~\ref{fig:gsbox} shows that the median values are largely similar, and the spread of accuracies is less variable (standard deviation is $\approx$ 0.02). Consequently, we conclude that the VR-B method with shortest-path distance (figure~\ref{fig:spdbox}) is quite robust against edge deletions (see Appendix figure~\ref{fig:graphresistencerobustness} for the graph resistance plot). 
Nevertheless, figure~\ref{fig:kernelbox} indicates discernible deviations in median values and the spread of accuracies for the Weisfeiler-Leman kernel method (standard deviation is $\approx$ 0.03). Given these deviations, it is essential to exercise caution when making claims about the robustness of the Weisfeiler-Leman method, as no clear trend suggesting monotonicity is observed. The outlier in the figure, represented by the low-value points around 0.3, corresponds to the $\texttt{ENZYMES}$ dataset. This finding indicates that perturbation significantly reduces the already low accuracy values for the $\texttt{ENZYMES}$ dataset.

\subsection{Interpretability of TDA}
Certain topological features hold the potential to provide valuable insights and explanations for the predictions made by machine learning models. We can identify significant topological features, such as connected components, loops, voids, and higher-dimensional topological structures.  

Topological features contribute to a model's prediction or output with varying degrees of importance. To gain insight into how different features (epsilon values) influence our results, we conducted an experiment using a Vietoris-Rips (VR) filtration. We computed the mean along the columns for each epsilon value, ranging from $\epsilon_1$ to $\epsilon_{100}$, and plotted the distribution for 0-, 1-, and 2- dimensional features as Betti functions. In Appendix figure~\ref{fig:bettidistribution}, we observe right-skewed distribution plots for each dataset, indicating that most features are born and die early in the filtration process. There is rarely any useful TDA signal in graphs after the filtration threshold $\epsilon_{60}$.

\begin{figure*}[h!]
\centering
\subfloat[1-dimensional topological features.]{\includegraphics[width=0.45\textwidth]{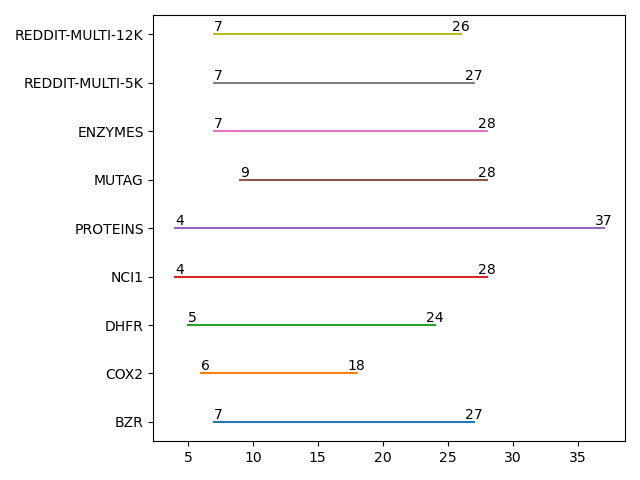}%
     \label{fig:b1}}
\hfil
\subfloat[VR complexity and Betti function.]{\includegraphics[width=0.5\textwidth]{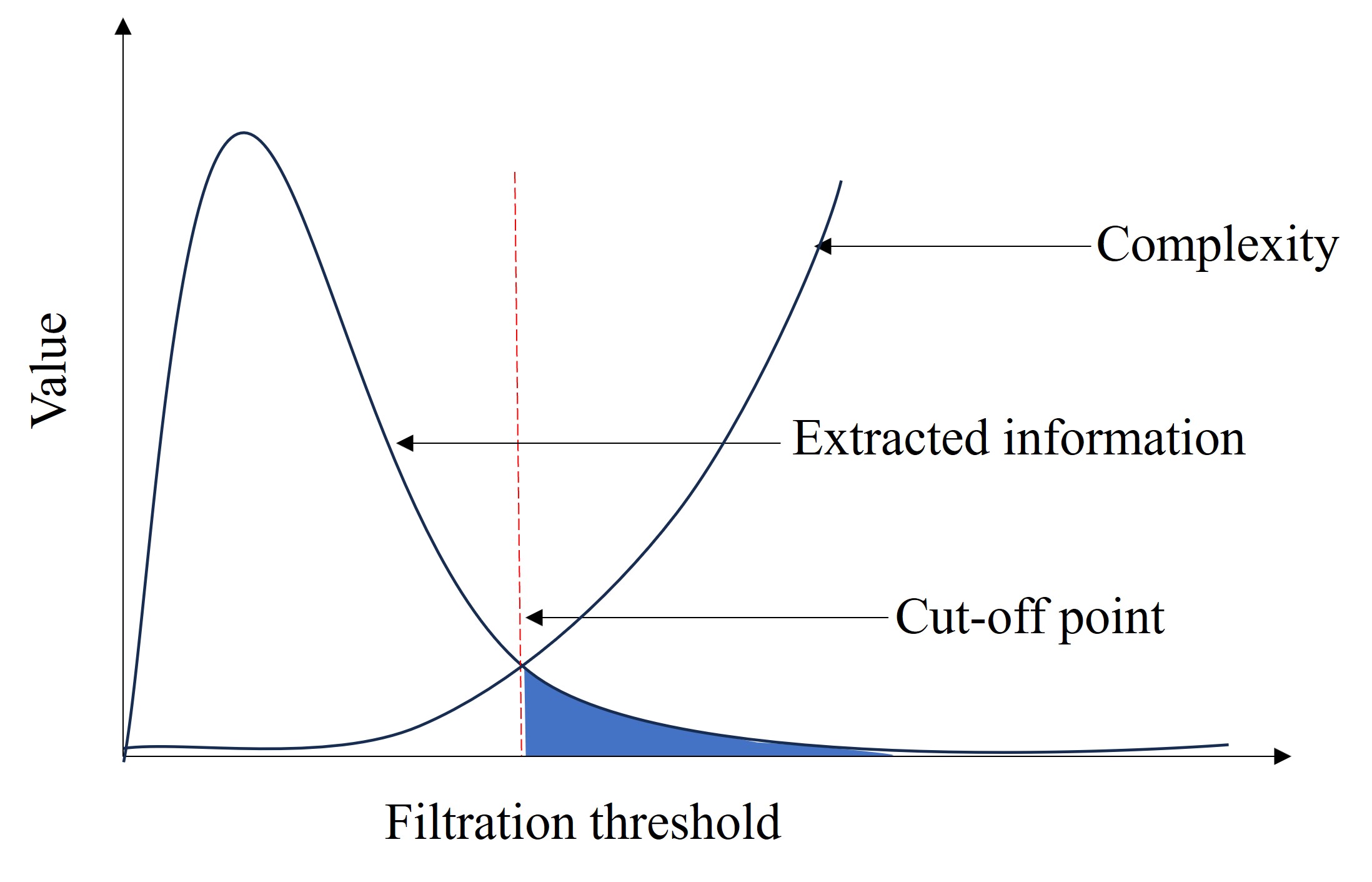}%
     \label{fig:complexityusefulness}}
\caption{Useful Filtration Ranges. The x-axis Values are the Filtration Thresholds (in $[0,100]$) Defined Over Shortest Path Distance Between Node Pairs, and a Longer Range Between the Minimum and Maximum Values Indicates Greater Variability and Importance of these Thresholds in Influencing the Model's Predictions. The Range in figure~\ref{fig:b1} is Between the Minimum and Maximum Values of the 10 Most Important Filtration Thresholds identified through the Shapley analysis.}
\label{fig:impstep}
\end{figure*}

In figure~\ref{fig:b1}, we conduct a detailed analysis of the TDA features by plotting the epsilon indices of the 10 most important features obtained through an experiment using Shapley values~\cite{lundberg2017unified} for each dataset in our classification task (see Appendix figure~\ref{fig:b0} for the 0-dimension). As figure~\ref{fig:b1} shows most datasets exhibit relatively short useful filtration ranges, i.e., $[4-28]$. Notably, for 1-dimensional features, the highest important threshold is 37 for the $\texttt{PROTEINS}$ dataset. These results provide insights into the filtration thresholds that significantly impact the model's predictive capabilities and highlight the variation in importance across different datasets.

The insights from figure~\ref{fig:b1} offer a potential solution to the computational costs associated with TDA models. In figure~\ref{fig:complexityusefulness}, we illustrate the relationship between complexity and useful TDA information using a Betti function across the filtration threshold. The plot shows that increasing complexity doesn't necessarily result in a proportionate gain in useful information during filtration (see Appendix figure~\ref{fig:bettidistribution} for Bettis of all 0, 1 and 2 dimensions). Building on this insight, we propose implementing a cut-off point in the filtration process to enhance efficiency without significantly sacrificing vital information. By identifying and retaining the most impactful filtration thresholds, we can streamline TDA analysis, making it more feasible and computationally efficient while preserving essential topological features that significantly influence predictions. Moreover, datasets can be preprocessed to identify the presence of topological features before investing resources in their extraction. Moving in this direction, we introduce a surrogate model in the next section to identify and assess the significance of topological features, enabling informed decisions about resource allocation for feature extraction. This proactive approach can lead to more efficient data analysis and resource utilization in TDA-based models.
 
\section{Topological Signals in Graph Datasets}\label{sec:surrogatemodel}
\noindent We outline three key questions to illuminate the predictive power of TDA for graph datasets: i) Can we predict the presence of topological features in graphs, as an alternative to resource-intensive TDA computations? ii) Which traditional graph features does TDA excel at capturing, is there such a correspondence? iii) Can TDA aid in discovering and visualizing relationships between graph datasets?

\begin{figure*}[!t]
\centering
\includegraphics[width=\textwidth]{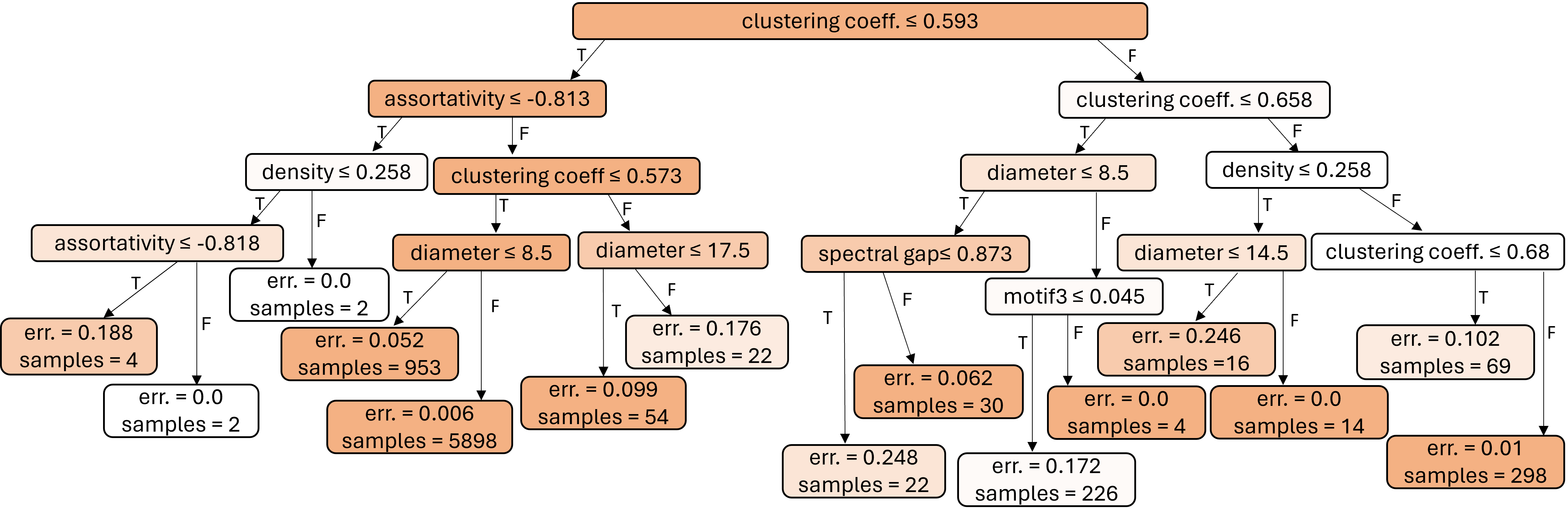}
\caption{Decision Tree Showing Features Split Using Decision Tree Regressor When the Target is Betti 1.} 
\label{fig:decisiontree}
\end{figure*}

\noindent{\textbf{Origins of 1D and 2D holes in graphs.}}  We build two predictive models over the graph features to study the existence of topological signals. Specifically, we use a Random Forest to analyze how changes in the graph features (e.g., higher clustering coefficient values) impact the presence of 1- and 2- dimensional topological features, providing insights that might not be immediately apparent from the raw data. The models utilize graph features as input and the counts of unique 1-dimensional (Betti 1) and 2-dimensional (Betti 2) features (along any filtration threshold) as the resulting outcomes in a Decision Tree (figure~\ref{fig:decisiontree}). We do not build a model for 0-dimensional features because in Vietoris-Rips, when the filtration starts every graph node is represented as a zero-dimensional hole. In other words, a naive predictor can output the number of nodes as Betti 0.  On the contrary, 1 and 2-dimensional features emerge in the filtration process at a subsequent stage.  

The performance metrics are laid out in Table~\ref{tab:regressionresult}, displaying an \textit{R2\_score} of $0.816$ for B1 and $0.7$ for B2. These figures signify that approximately $81.6$\% and $70$\% of the variations in the dependent variable (labels) can be comprehended through the Random Forest model. 

It is essential to note that the true value of the predictive models lies in deciding whether any n-dimensional hole exists within a graph. We use the models to determine whether resource-intensive TDA computations are warranted, offering valuable insights to reduce TDA's run-time cost. To illustrate, consider the well-known graph datasets IMDB-Binary~\cite{cai2018simple} and IMDB-Multi~\cite{zhao2019learning}, where the graphs exhibit dense connectivity and as a result lack $n>0$-dimensional features. Existing TDA libraries take days to arrive at the conclusion that 1, 2 or any other higher dimensional features do not exist in these graphs. 

Additionally, we employ the Shapley~\cite{lundberg2017unified} method to conduct an in-depth analysis of how graph attributes contribute to the predictive outcomes of our model. The application of Shapley analysis proves invaluable in comprehending the precise roles of individual features in steering predictive outcomes, revealing intricate interactions and their consequences.  

\begin{figure*}[!t]
    \centering
    \subfloat[1-dimensional features]{\includegraphics[width=2.5in]{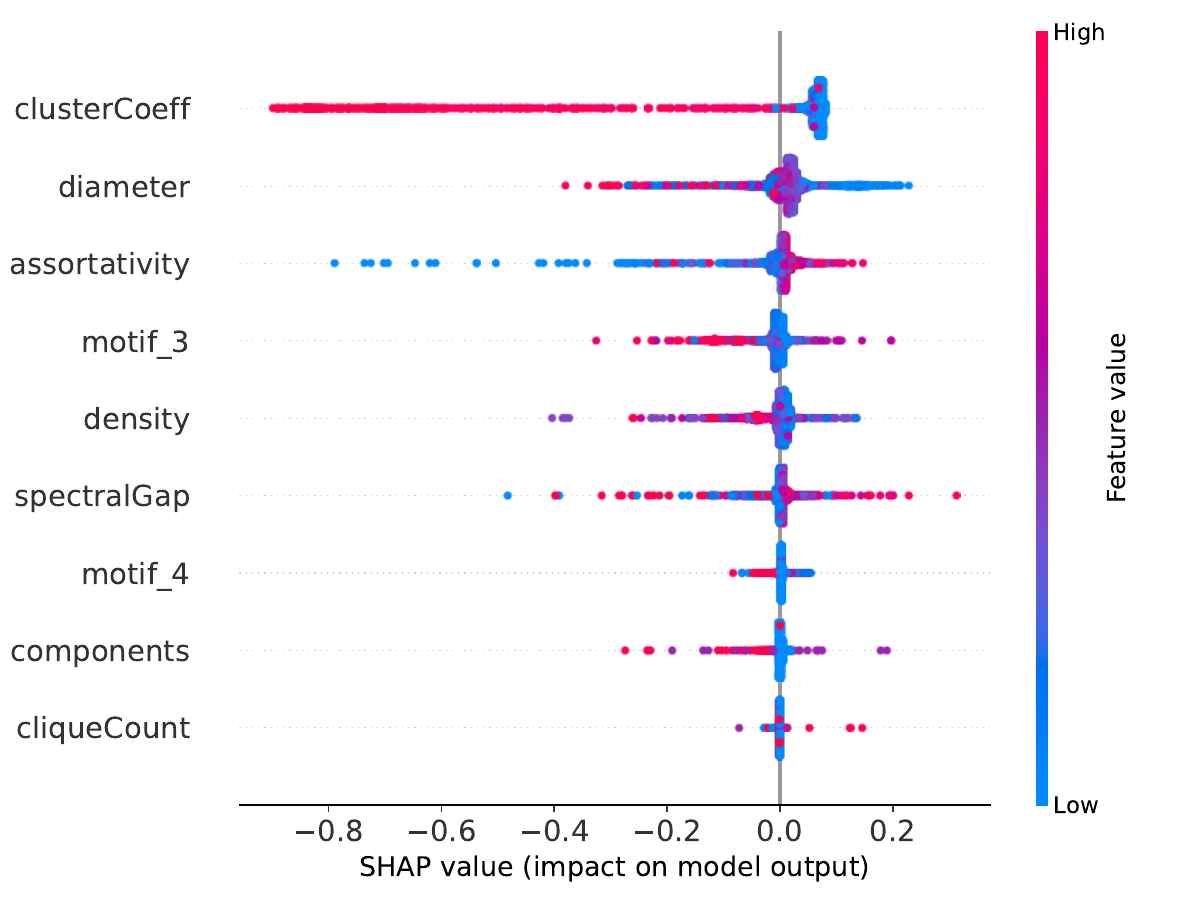}%
         \label{fig:shapleyb1}}
    \hfil
     \subfloat[2-dimensional features]{\includegraphics[width=2.5in]{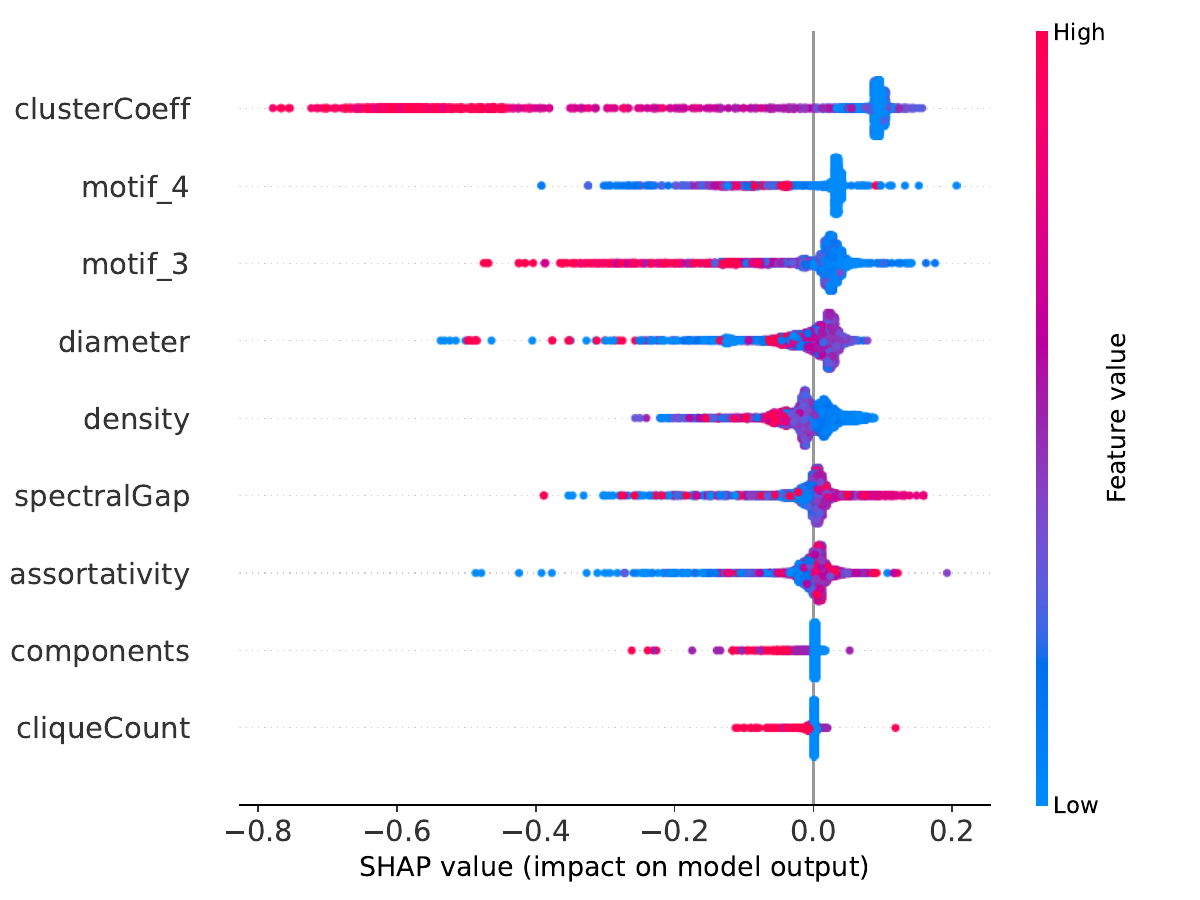}%
         \label{fig:shapleyb2}}
    \caption{Distribution Plot Illustrating the Importance of Each Feature in Predicting Betti 1 and Betti 2 Values. The x-axis Represents the Shapley Values, while the y-axis Represents the Features. Features are Color-Coded: Red Indicates High Values, Blue Signifies Low Values. Red Dots on the Negative Axis Indicate a Negative Impact of High Feature Values on Predictions, while Blue Dots on the Negative Axis Signify a Negative Impact of Low Feature Values. The Interpretation is Analogous for Red and Blue Dots on the Positive Axis.}
    \label{fig:surrogateModel}
\end{figure*}

As showcased in figure~\ref{fig:surrogateModel}, the clustering coefficient takes center stage for both B1 and B2. This implies that heightened clustering coefficients exert a detrimental impact on the existence of 1-dimensional and 2-dimensional characteristics. Assortativity and diameter also play pivotal roles in predicting Betti 1. Moreover, Motif 3 (referred to as $\angle$, representing the count of unclosed triangles) and spectral gap are incorporated into the analysis, albeit with relatively lesser significance compared to the aforementioned attributes.

\begin{table}[hbt!]
\centering
\caption{{Regression metrics results for the two surrogate models to predict Betti 1 (B1) and Betti 2 (B2) values.} \label{tab:regressionresult}}
\begin{tabular}{lll}
\toprule
Metric & B1 & B2\\
\midrule
MAE & 0.035 & 0.086\\
RMSE & 0.127 & 0.220\\
R2-SCORE $\uparrow $ & 0.816 & 0.700\\
\bottomrule
\end{tabular}
\end{table}

\noindent{\bf{Surrogate model to visualize TDA.}} A surrogate model~\cite{du2019techniques} is a valuable analytical tool utilized in various fields to approximate and comprehend complex relationships between input variables and corresponding outcomes. The model serves as a simplified representation of the original system, facilitating easier interpretation and analysis. In our setting, we create a binary decision tree surrogate to explain the TDA signals. To be specific, the model employs graph attributes as its input, while the resulting outcomes consist of the binarized tallies of distinct 1-dimensional (Betti 1) and 2-dimensional (Betti 2) characteristics, irrespective of the filtration threshold.

A decision tree is inherently explainable; it illuminates which features increase or decrease the likelihood of discovering topological signals in graphs. We find these feature divisions to be visually informative. As illustrated in figure~\ref{fig:decisiontree}, the decision tree, built on graphs from all datasets, identifies the clustering coefficient as a pivotal feature (see Appendix figure~\ref{fig:decisiontreeb2} for the tree of B2). Graphs characterized by high clustering coefficients tend to exhibit fewer voids. \cite{coraltdaAkcora} had conjectured that \textquote{when the clustering coefficient of a graph is either too low or too high, the higher-order persistence diagrams associated with this data tend to be trivial}. Their empirical results revealed that the ceiling of the coefficient was $\beta_k \approx 0.7$ (figure 10 of~\cite{coraltdaAkcora}). In line with their findings, our surrogate model (figure~\ref{fig:decisiontree}) identifies the highest cutoff value as $\beta_k \approx 0.68$.

Higher density values correspond to a reduction in the number of gaps within the graph as well. The significance of assortativity is notable, primarily due to its role in the creation of one-dimensional gaps in chain-like graphs (e.g., $v_1\rightarrow v_2\rightarrow v_3 \rightarrow v_4 \rightarrow v_1$), where nodes with comparable degrees (such as $v_2$ and $v_3$) are sequentially connected. This sequential linkage leads to positive assortativity, contributing to the formation of these one-dimensional voids.

\noindent{\textbf{Explainability: Visualizing graph datasets.}} 
In our predictive models we carefully filtered out irrelevant graph features to ensure the surrogate model's efficacy. As a result, we retain only the pertinent features, which we then utilize within the framework of TDA Mapper~\cite{carlsson2021topological}. TDA Mapper operates by constructing a network of clusters, wherein each node represents a cluster of graphs based on their N-dimensional features, and edges indicate shared graphs. This approach facilitates the identification of structural relationships among the datasets. Notably, proximity within the network signifies similar feature characteristics. 

Interestingly, our findings in figure~\ref{Mapper} revealed certain expected patterns, such as the similarity among \texttt{REDDIT} datasets (they come from the same domain). Moreover, we observed a  similarity between $\texttt{PROTEINS}$ and $\texttt{ENZYMES}$ datasets as the enzymes are themselves a subset of proteins.  Some of the $\texttt{NCI1}$ graphs appear in the two small disconnected components. Most importantly, the Mapper network shows that we can group the datasets into four: 1 - $\texttt{ENZYMES}$,  $\texttt{PROTEINS}$, 2-  $\texttt{REDDIT-MULTI-5K}$,  $\texttt{REDDIT-MULTI-12K}$, 3-  $\texttt{NCI1}$,  $\texttt{MUTAG}$, 4-  $\texttt{BZR}$,  $\texttt{COX2}$,  $\texttt{DHFR}$. 

It is advisable to conduct GraphML experiments in light of these findings. It is important to consider that similar datasets might not offer substantial complementary evidence when evaluating model performance.

\section{Conclusion} \label{sec:conclusion}
\noindent In this work, we have carried out experiments to explain the power, complementary power, scalability, robustness, and interpretability of topological data analysis in the context of graph classification. Our observations indicate that TDA does not outperform graph kernels in terms of model accuracy; however, it can provide supplementary information for utilization. Moreover, TDA enhances robustness against graph perturbations. The viability of TDA depends on its scalability with large graphs, which remains limited. Nevertheless, we identify filtration thresholds where topological signals can be extracted efficiently. These insights offer the potential to scale TDA for use with larger graphs.


\bibliographystyle{IEEEtran}
\bibliography{IEEEabrv, xtda} 

\clearpage

\appendices

\noindent Appendix sections give additional details for our experiments and methods. In \ref{sec:tutorial}, we give details on persistence-based topological methods, including persistent homology in GraphML, node distance in GraphML, persistence diagrames in GraphML and their vectorizations. In \ref{sec:graphFeatures}, we discuss features extracted from graphs for our classification and in \ref{sec:apxkernel}, we discuss the graph kernel methods adopted. The statistics of the graphs used are given in \ref{AppendixA}, model hyperparameters in \ref{sec:modelhyper}. Further details on our research questions and additional results are given from \ref{sec:graphFeatures} to \ref{sec:mapper} and finally, related works are discussed in \ref{sec:relatedwork}.

\section{Feature-based Graph Classification}  \label{sec:graphFeatures}
\noindent We extract the following features for each graph in a dataset to carry out a basic classification task which will serve as a baseline for other methods. 

\begin{itemize}
\item Graph density: Graph density measures the proportion of actual connections present in a graph compared to the total possible connections. It indicates how dense or sparse the connections are in the graph.

\item Graph diameter: The graph diameter is the longest shortest path length between any two nodes in a graph. It represents the maximum distance between any pair of nodes and provides an insight into the overall size of the graph.

\item Clustering coefficient: The clustering coefficient measures the extent to which nodes in a graph tend to form clusters or groups. It quantifies the likelihood that the neighbors of a node are also connected to each other, indicating the presence of local clustering.

\item Spectral gap: The spectral gap refers to the difference between the largest and the second-largest eigenvalue in the graph's adjacency matrix. It provides information about the connectivity and expansion properties of the graph.

\item Node assortativity: Node assortativity measures the tendency of nodes in a graph to connect with similar nodes. It quantifies whether nodes with high degrees tend to connect with other nodes with high degrees (assortative mixing) or with nodes with low degrees (disassortative mixing).

\item Number of cliques in the graph: A clique is a subset of nodes in a graph where every node is directly connected to every other node. The number of cliques in a graph indicates the presence of tightly interconnected subgroups.

\item Number of disconnected components: Disconnected components refer to groups of nodes in a graph that are not connected to each other. The number of disconnected components provides insights into the graph's overall connectivity and potential for isolated subgraphs.

\item Three-vertex motif counts: Three-vertex motifs are small patterns formed by three interconnected nodes in a graph. Counting the number of different three-vertex motifs in a graph can reveal specific patterns or motifs that occur frequently, which may have implications for network dynamics or functionality.
\end{itemize}

Henceforth, we will collectively refer to these characteristics as \textit{graph features} in the remaining sections of the paper.

\noindent\textbf{Preprocessing graph features:} It is important to note that certain graph features have the potential to leak dataset-specific information which may bias our results. As an example, RED-12K graphs display the largest size and motif counts, which means that we may inadvertently associate a large graph size with the topological characteristics of RED-12K graphs. To mitigate this issue, we max-normalize three-node motif counts by $\left | \V \right |\times (\left | \V \right |-1) \times (\left | \V \right |-2)/6$ to prevent any inadvertent data leakage. We do not normalize diameters because large graphs may have small diameters as stated in~\cite{leskovec2005graphs}.

\section{Graph Kernel Methods}\label{sec:apxkernel}
\noindent\textit{Weisfeiler-Leman subtree kernel}~ introduced by~\cite{shervashidze2011weisfeiler} is based on the idea of iteratively propagating (node) label information through a graph and aggregating labels from neighbors, which results in a feature vector representation that can be used to assess the dissimilarity of graphs~\cite{rieck2019persistent}. Our choice of focusing on WL is due to an important result in graph classification, which states that popular message passing graph neural networks cannot distinguish between graphs that are indistinguishable by the 1-WL test~\cite{morris2019weisfeiler}. For this reason, we use the WL kernel as a baseline of what message passing GNNs can achieve and compare its performance to the TDA approaches. In addition, a base kernel for graphs is needed for the computation of the WL kernel in practice. The base kernel refers to the initial kernel matrix that is computed based on the node labels of the original graph. This kernel matrix serves as the starting point for the recursive hashing procedure.

\begin{definition}[\textbf{Weisfeiler-Leman Kernel}]
Suppose $k$ is any kernel for graphs, say $U$ and $U^{\prime}$, we call $k$ the base kernel. Then, the WL kernel with $h$ iterations and base kernel $k$ is defined as $$k_{WL}^{(h)}(U,U^{\prime}) = k(U_0,U_0^{\prime}) + k(U_1,U_1^{\prime})+\dots+k(U_h,U_h^{\prime}),$$ where $h$ is the number of WL iterations and $\{U_0,\dots,U_h \}$ and $\{U_0^{\prime},\dots,U_h^{\prime}\}$ are the WL sequences of $U$ and $U^{\prime}$ respectively~\cite{shervashidze2011weisfeiler}.
\end{definition}

The graphs $U_i$ and $U_i^{\prime}$ for $i={0,\dots,h},$ are the relabelled graphs. This scheme is expected to terminate (in general) after at most $max\{|V(U)|, |V(U^{\prime})|\}$ iterations~\cite{morris2021power}. For more on the WL kernel, we refer our readers to~\cite{shervashidze2011weisfeiler}. In our work, we use VertexHistogram\footnote{The vertex histogram constructs histograms based on the vertex attributes where histogram bins represent attribute value ranges, and the bin values indicate the frequency or count of vertices falling into each bin.} as the base kernel and $h\in [2,3]$ iterations. We then transform the test graphs into the same feature space as the training graphs and use the resulting kernel matrix say $\mathcal{P}$, as a feature vector for our graph classification task.

\section{Statistics of the benchmark graph datasets.}\label{AppendixA}

\begin{table}[hbt!]
\centering 
\begin{tabular}{lcccc}
\toprule
\multicolumn{1}{c}{Datasets}&\multicolumn{1}{c}{Graphs\# }&\multicolumn{1}{c}{Average\_nodes\#}&\multicolumn{1}{c}{Average\_edges\#}&\multicolumn{1}{c}{Classes}\\ \midrule
$\texttt{BZR}$&$405$&$35.75$&$38.36$&$2$\\
$\texttt{COX2}$&$467$&$41.22$&$43.45$&$2$\tabularnewline
$\texttt{DHFR}$&$756$&$42.43$&$44.54$&$2$\tabularnewline
$\texttt{ENZYMES}$&$600$&$32.63$&$62.14$&$6$\tabularnewline
$\texttt{MUTAG}$&$188$ & $4.23$&$19.79$&$2$\tabularnewline
$\texttt{NCI1}$&$4110$&$29.87$&$32.30$&$2$\tabularnewline
$\texttt{PROTEINS}$&$1113$&$39.06$&$72.82$&$2$\tabularnewline
$\texttt{REDDIT-5K}$&$4999$&$508.52$&$594.87$&$5$\tabularnewline
$\texttt{REDDIT-12K}$&$11929$&$391.41$&$456.89$&$11$\tabularnewline
\bottomrule
\end{tabular}
\end{table}
$\texttt{BZR}$, $\texttt{COX2}$ and $\texttt{DHFR}$ are all chemical compound datasets for which the nodes in each graph represent atoms and edges represent chemical bonds. The task is to classify compounds as active or inactive~\cite{sutherland2003spline}.
$\texttt{NCI1}$ represent a balanced subset of datasets of chemical compounds screened for activity against non-small cell lung cancer and ovarian cancer cell lines respectively~\cite{wale2008comparison}. $\texttt{PROTEINS}$ is a graph model of proteins introduced by~\cite{borgwardt2005protein}. For this dataset, the nodes represent secondary structure elements and an edge exists if two nodes are neighbours along the amino acid sequence or one of the three nearest neighbours in space. The task for this dataset is to predict whether a protein is an enzyme~\cite{morris2020tudataset}.  $\texttt{ENZYMES}$ is a dataset of protein tertiary structures obtained from~\cite{borgwardt2005protein}, consisting of 600 enzymes from the BRENDA enzyme database~\cite{schomburg2004brenda}. In this case, the task is to correctly assign each enzyme to one of the 6EC top-level classes which reflect the catalyzed chemical reaction~\cite{morris2020tudataset}. $\texttt{MUTAG}$ is a graph structured dataset consisting of 188 chemical compounds which is divided into two classes according to their mutagenic effect on a bacterium. The nodes represents atoms and edges represents chemical bonds~\cite{debnath1991structure}. The task here is to predict mutagenicity. $\texttt{REDDIT-5K}$ and $\texttt{REDDIT-12K}$~\cite{yanardag2015deep} are social networks graphs representing a discussion thread, where the nodes corresponds to users and an edge exist if one user responds to a comment of another user. The task is to predict the subreddit, where the thread was posted. $\texttt{REDDIT-5K}$ has five subreddits and $\texttt{REDDIT-12K}$ eleven subreddits.

\begin{figure*}[hbt!]
\centering
 \subfloat[Graph five from MUTAG dataset.]{
     \includegraphics[width=0.35\textwidth]{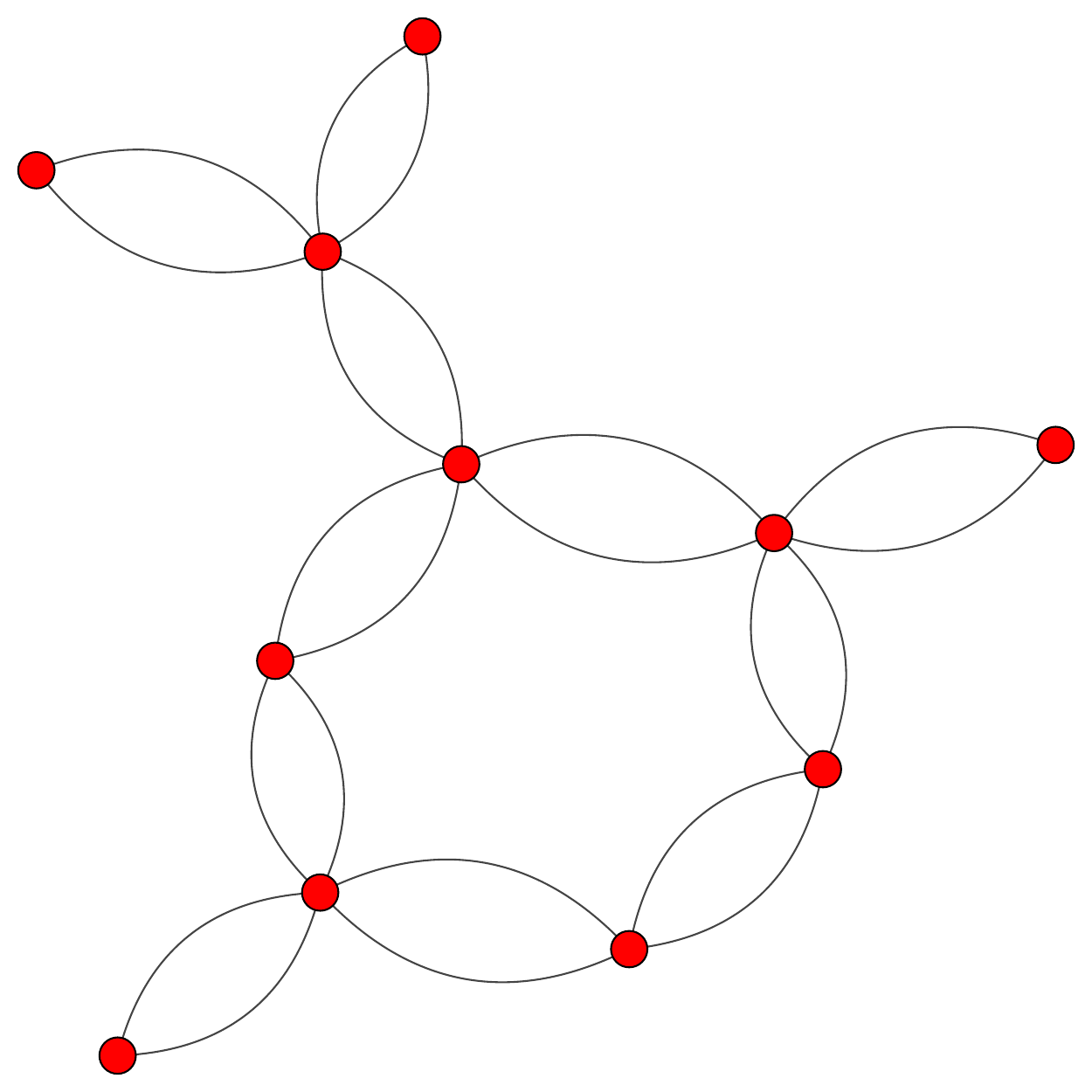}%
     \label{fig:mutaggraph}}
\hfil
\subfloat[The resulting Persistence Diagram of~\ref{fig:mutaggraph}. Betti numbers $\beta_0$ and $\beta_1$ are the count of red and blue circles, respectively.]{
     \includegraphics[width=0.35\textwidth]{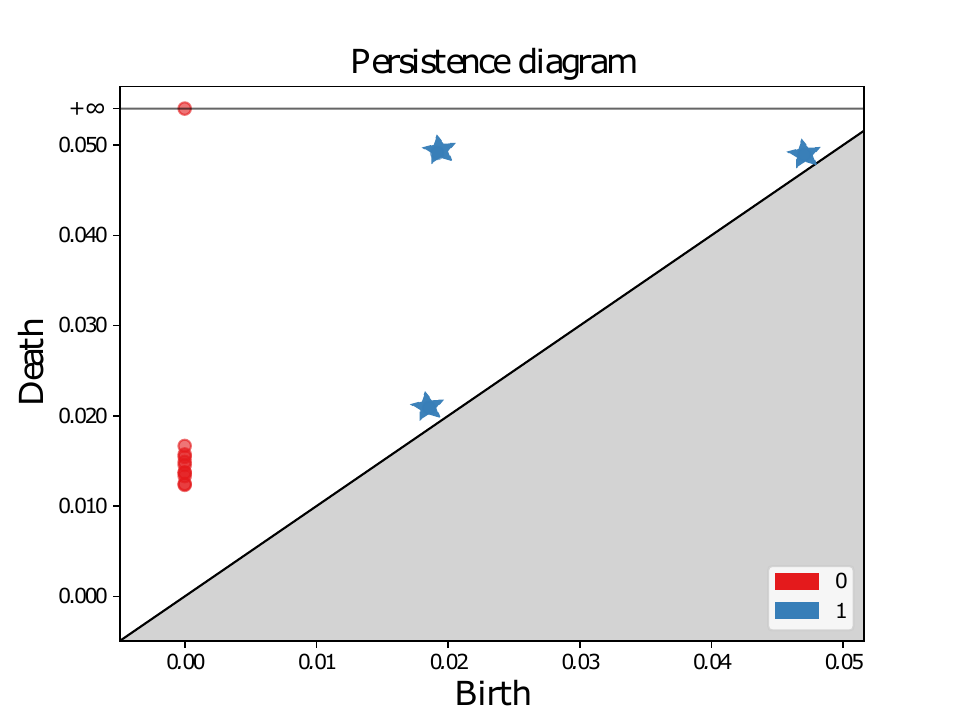}%
     \label{fig:mutagPD}}
\hfil
\subfloat[Persistence landscape of~\ref{fig:mutaggraph} for $k=1$.]{\includegraphics[width=0.35\textwidth]{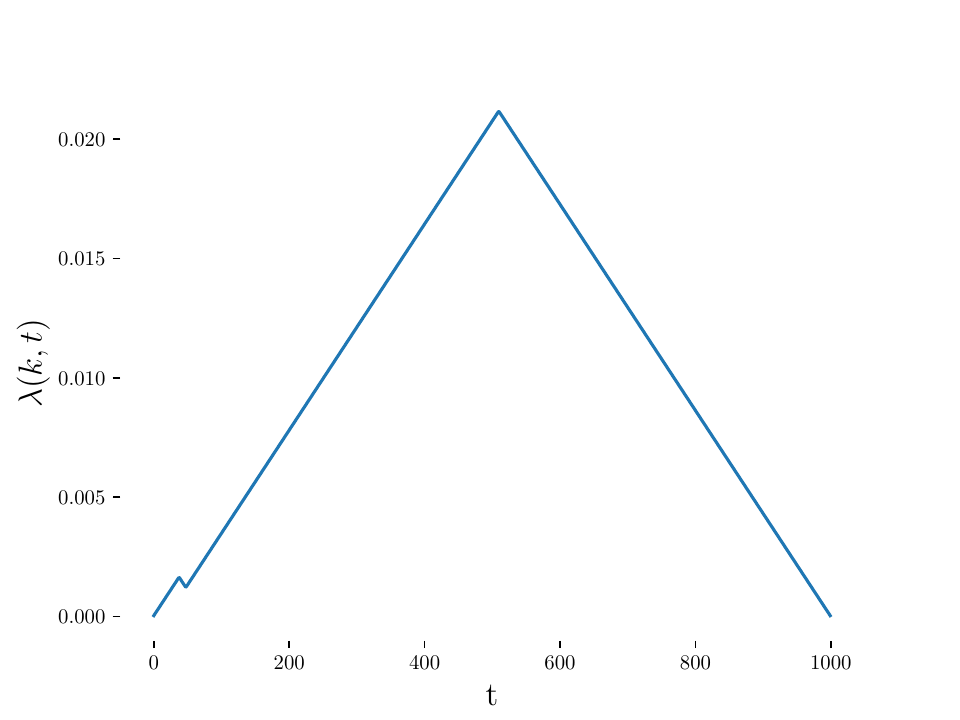}%
     \label{fig:mutagPL}}
\hfil
\subfloat[Persistence silhouette of~\ref{fig:mutaggraph} for $w=1$.]{\includegraphics[width=0.35\textwidth]{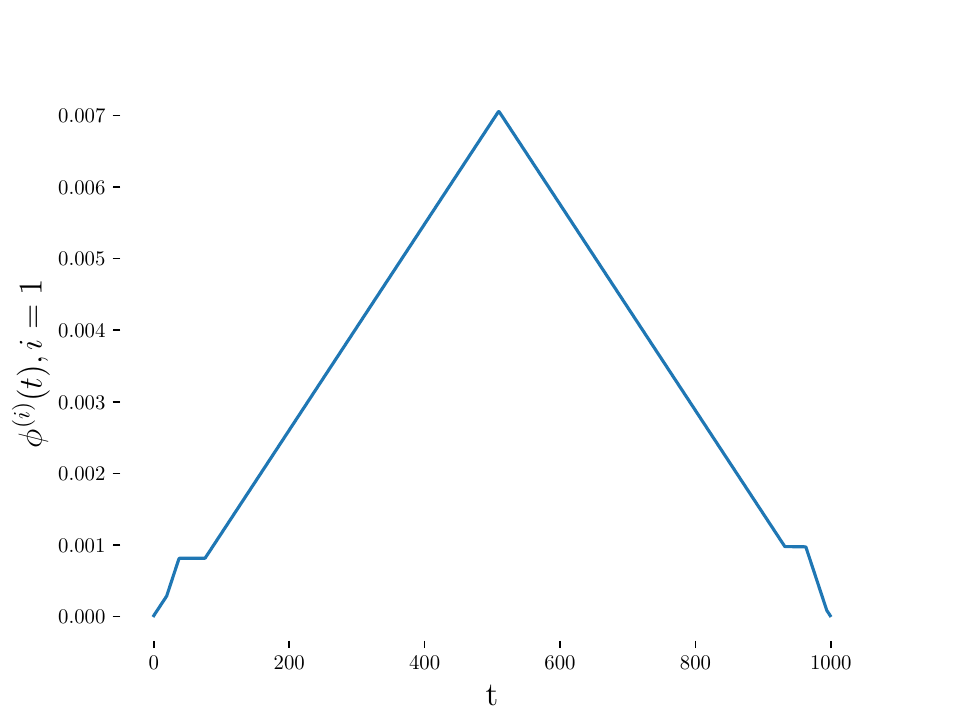}%
     \label{fig:mutagSilh}}
\caption{Pictorial representation of the different forms of vectorization for a sample graph from MUTAG data.}
\label{fig:vectorization}
\end{figure*}

\section{Model Hyperparameters} \label{sec:modelhyper}
\begin{enumerate}
    \item Alpha complex: {\fontfamily{qcr}\selectfont max-alpha-square} = {\fontfamily{qcr}\selectfont inf}.
    \item Vietoris-Rips complex: {\fontfamily{qcr}\selectfont thresh=1, maxdim=1}
    \item Weisfeiler-Leman~: {\fontfamily{qcr}\selectfont n-iter = [2, 3] and base-kernel-graph = VertexHistogram}
\end{enumerate}

\begin{table}[hbt!]
\centering
\begin{tabular}{ll}
\toprule
Methods & Meaning\\
\midrule
VR-B & Vietoris-Rips Betti\\
VR-L & Vectorizing VR using persistence landscape \\
VR-S & Vectorizing VR using persistence silhouette \\
Base + VR-B & Baseline + Vietoris-Rips Betti\\
AC-B & Alpha complex Betti \\
AC-L & Vectorizing AC using persistence landscape \\
AC-S & Vectorizing AC using persistence silhouette \\
Base + AC-B & Baseline + Alpha complex Betti \\
WL & Weisfeiler-Leman \\ 
Baseline & Graph characteristics model \\
\bottomrule
\end{tabular}
\end{table}

\begin{table}
\centering
\begin{tabular}{lll}
\toprule
Feature & Range & Normalization\\
\midrule
Density & 0 - 4 & - \\
Diameter & 1 - 64 & - \\
Clustering Coefficient & 0 - 1 & - \\
Spectral Gap & 0 - 12175  & - \\
Assortativity & -1 - 0.7 & - \\
Motif3 (2 edges) & 0 - 4688750 & $\left|V\right|$($\left|V\right|$-1)($\left|V\right|$-2)/6 \\
Motif4 (complete triangle) & 0 - 1599 & $\left|V\right|$($\left|V\right|$-1)($\left|V\right|$-2)/6 \\
No. of Component & 1 - 27 & - \\
Clique & 2 - 6 & - \\
\bottomrule
\end{tabular}
\end{table}

\begin{figure*}[hbt!]
\centering
\subfloat[BZR]{
     \includegraphics[width=0.3\textwidth]{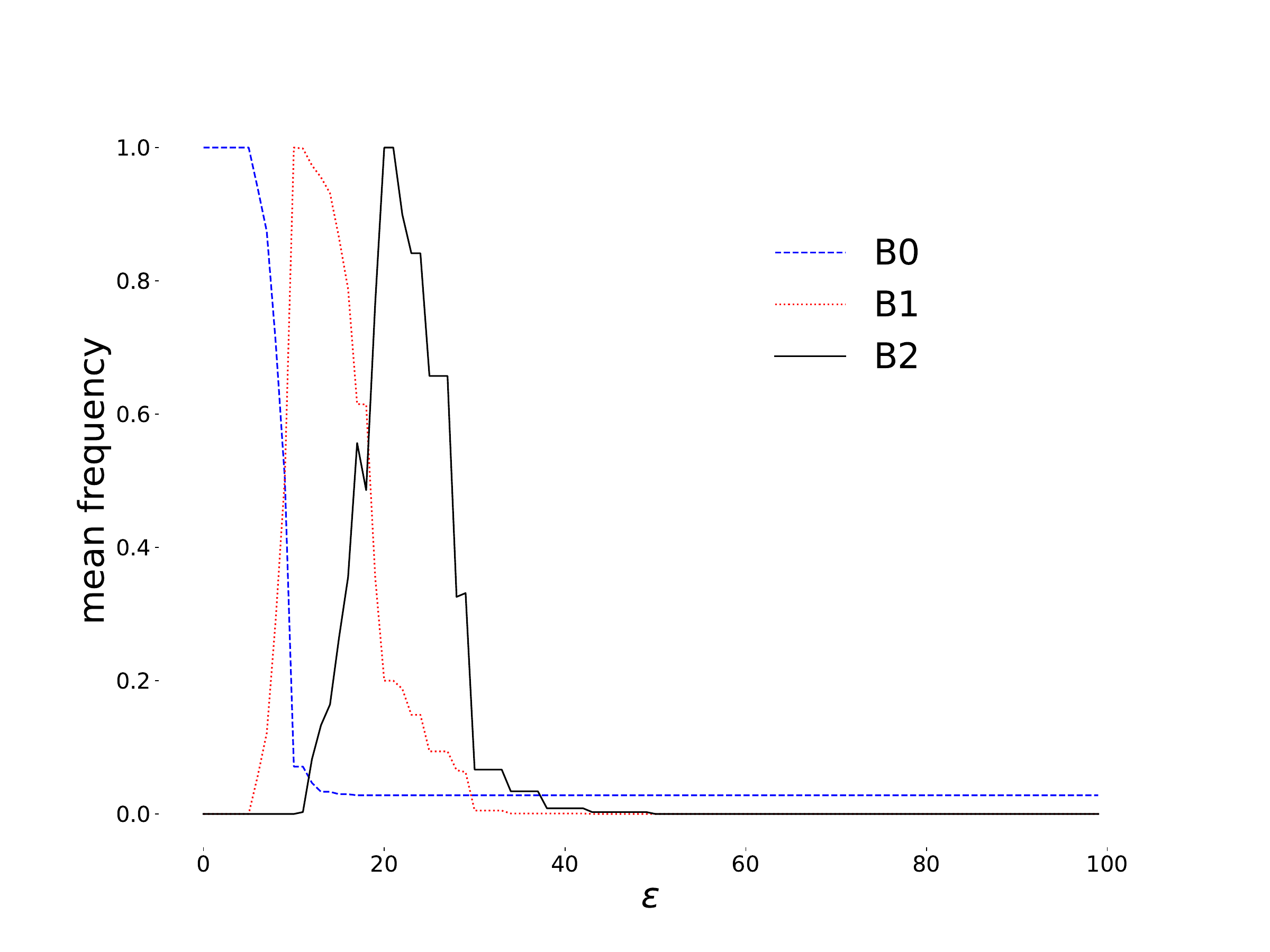}%
     \label{fig:aa}}
\hfil
\subfloat[COX2]{
     \includegraphics[width=0.3\textwidth]{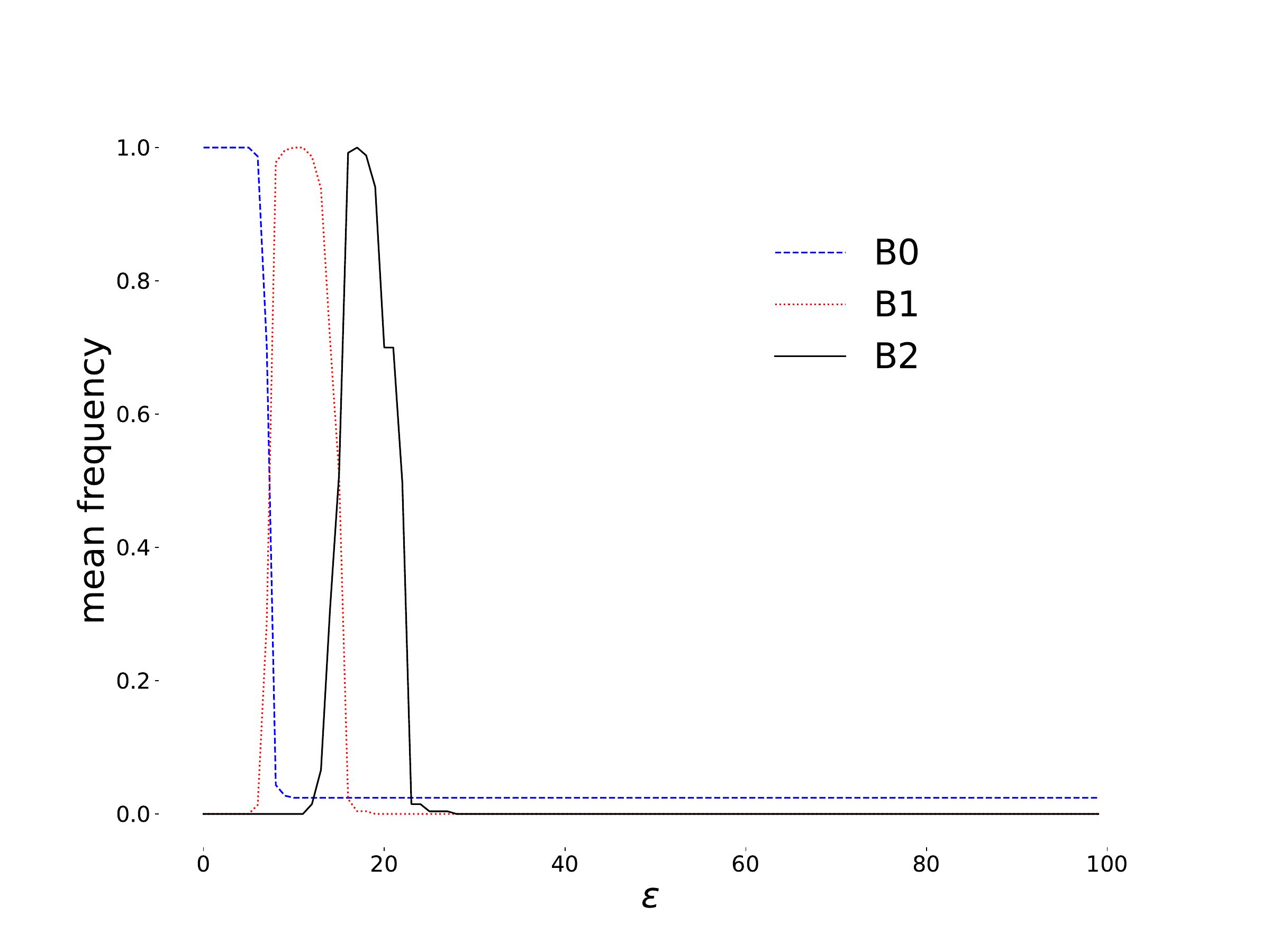}%
     \label{fig:ab}}
\hfil
\subfloat[DHFR]{
     \includegraphics[width=0.3\textwidth]{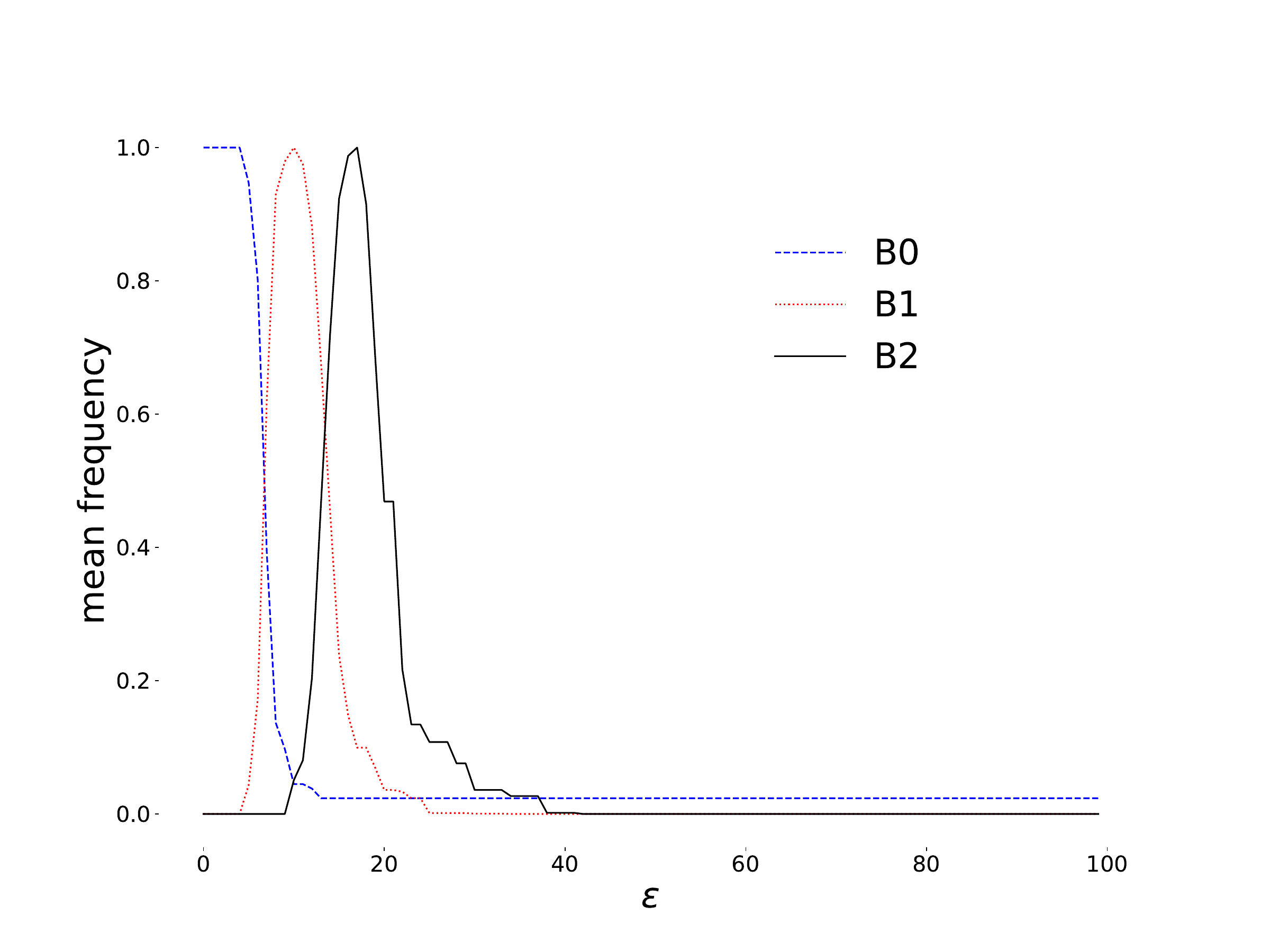}%
     \label{fig:ac}}
\hfil
\subfloat[ENZYMES]{
     \includegraphics[width=0.3\textwidth]{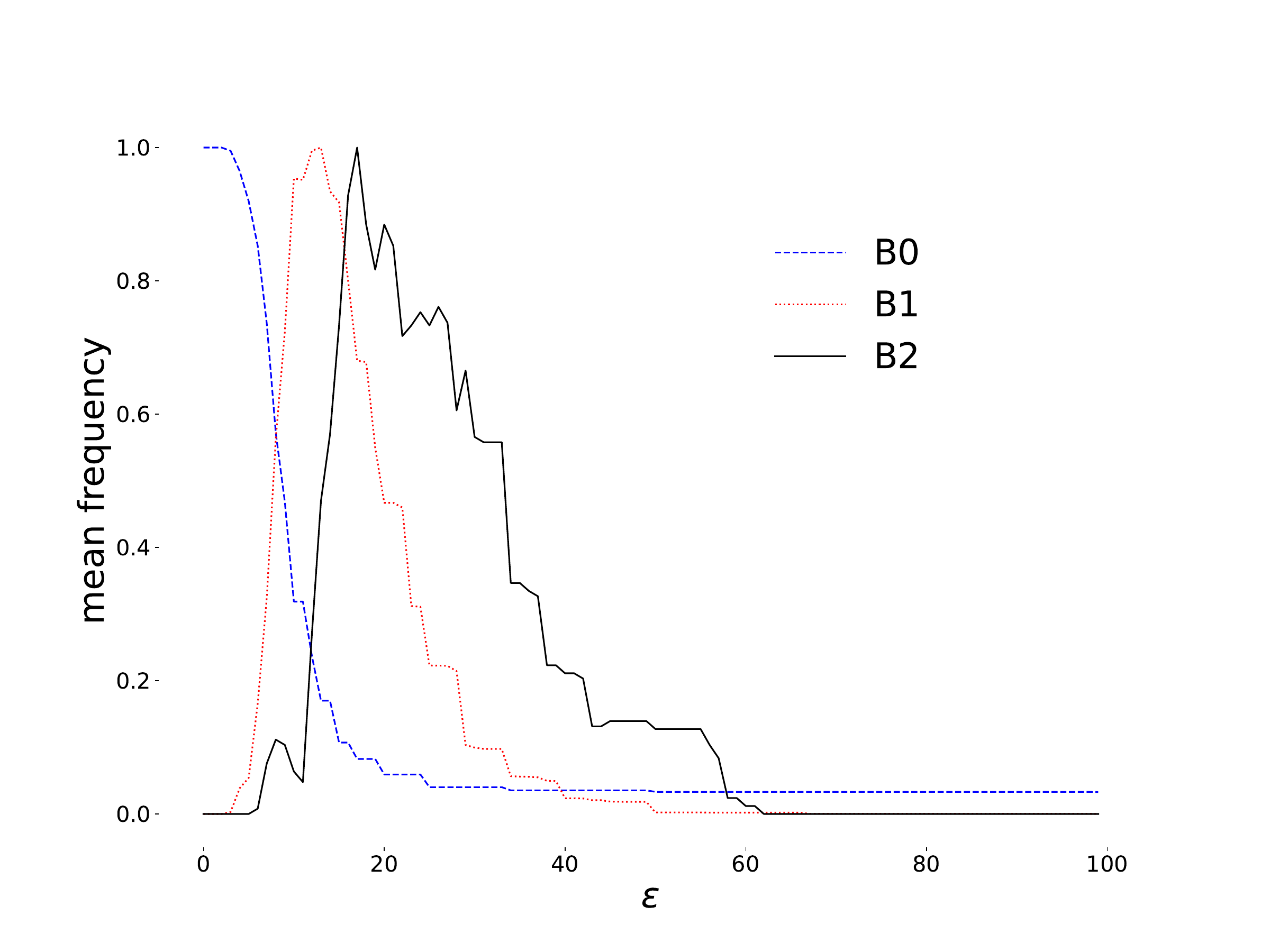}%
     \label{fig:ad}}
\hfil
\subfloat[MUTAG]{
     \includegraphics[width=0.3\textwidth]{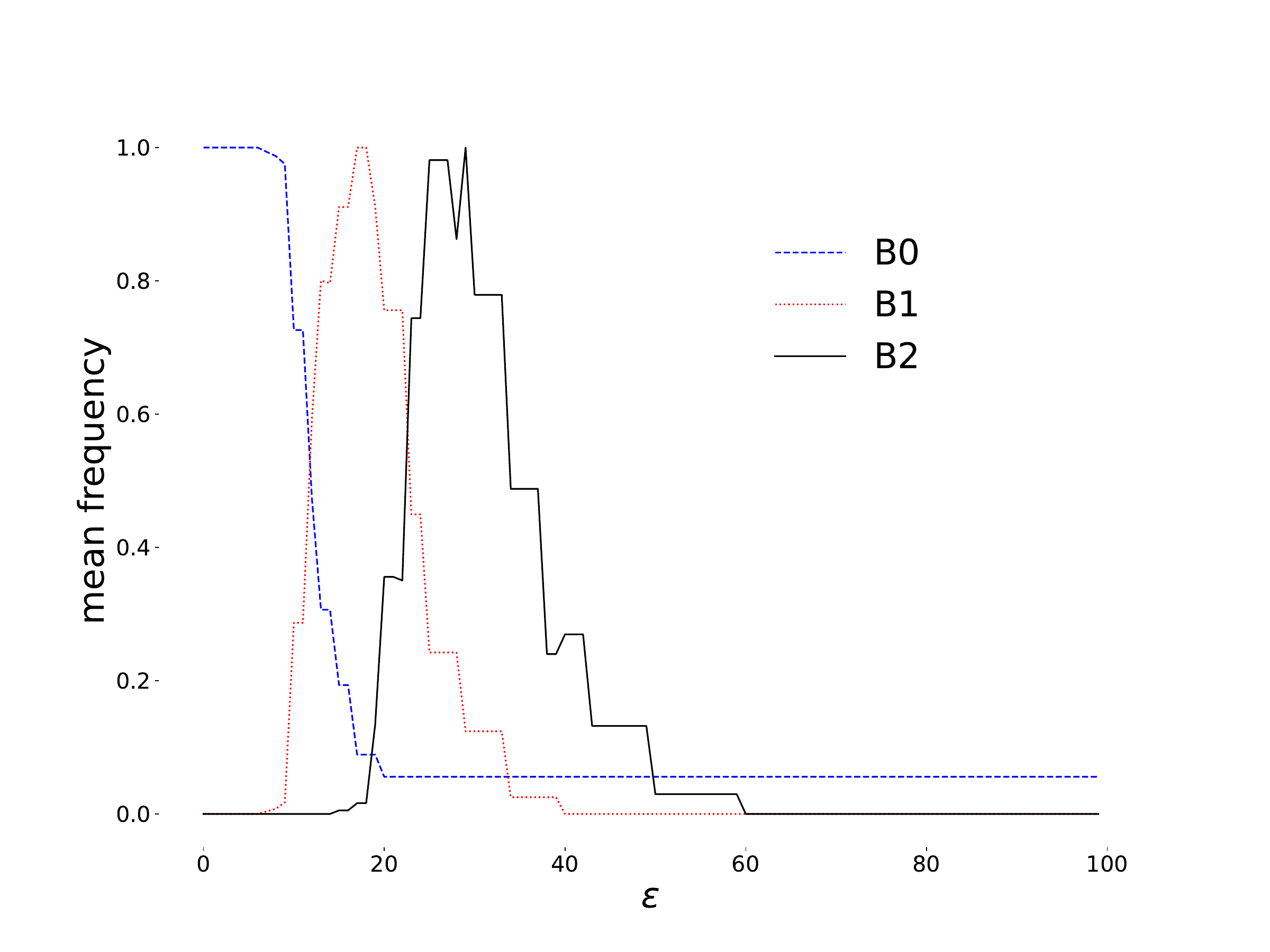}%
     \label{fig:ae}}
\hfil
\subfloat[NCI1]{
     \includegraphics[width=0.3\textwidth]{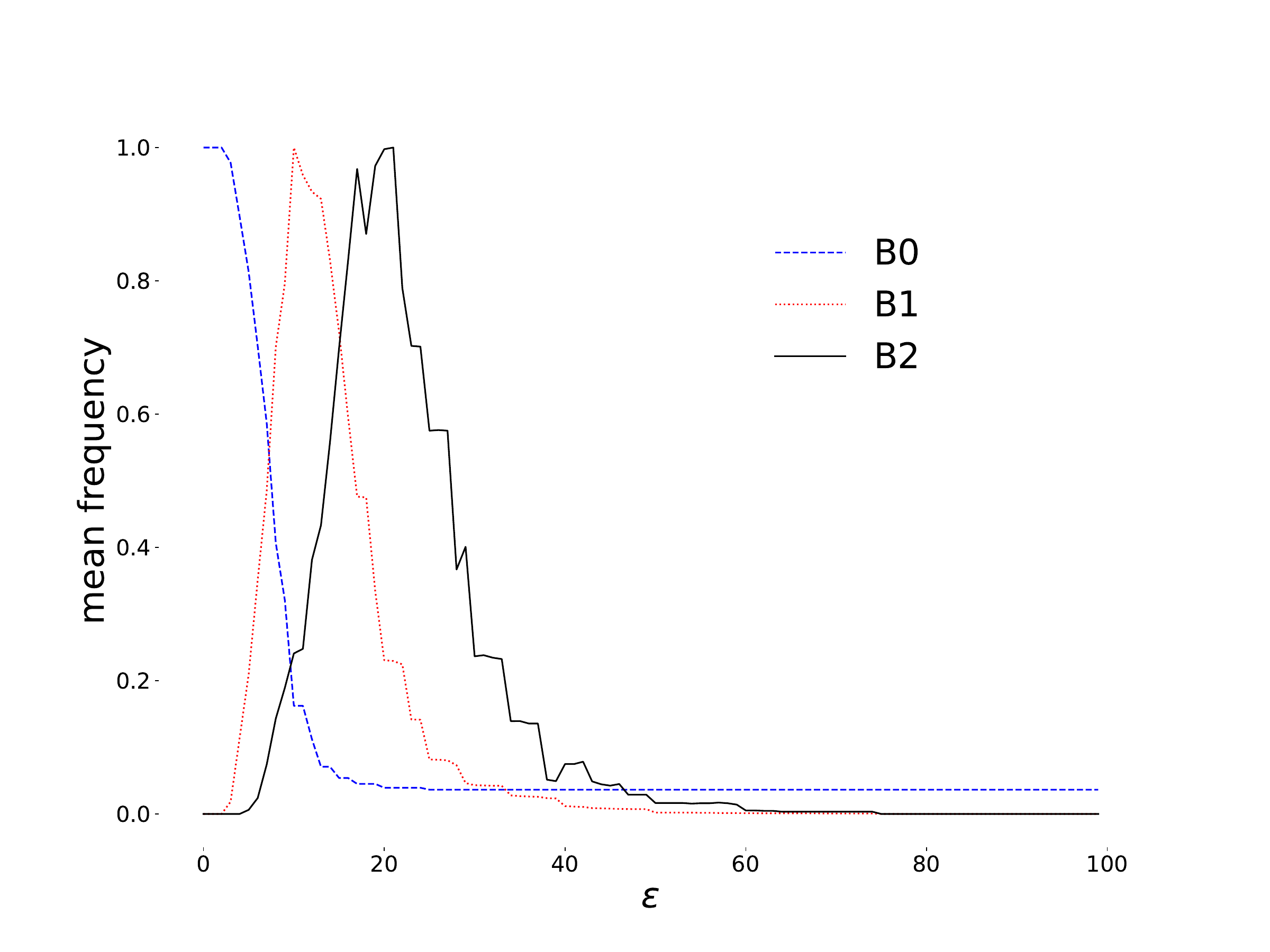}%
     \label{fig:af}}
\hfil
\subfloat[PROTEINS]{
     \includegraphics[width=0.3\textwidth]{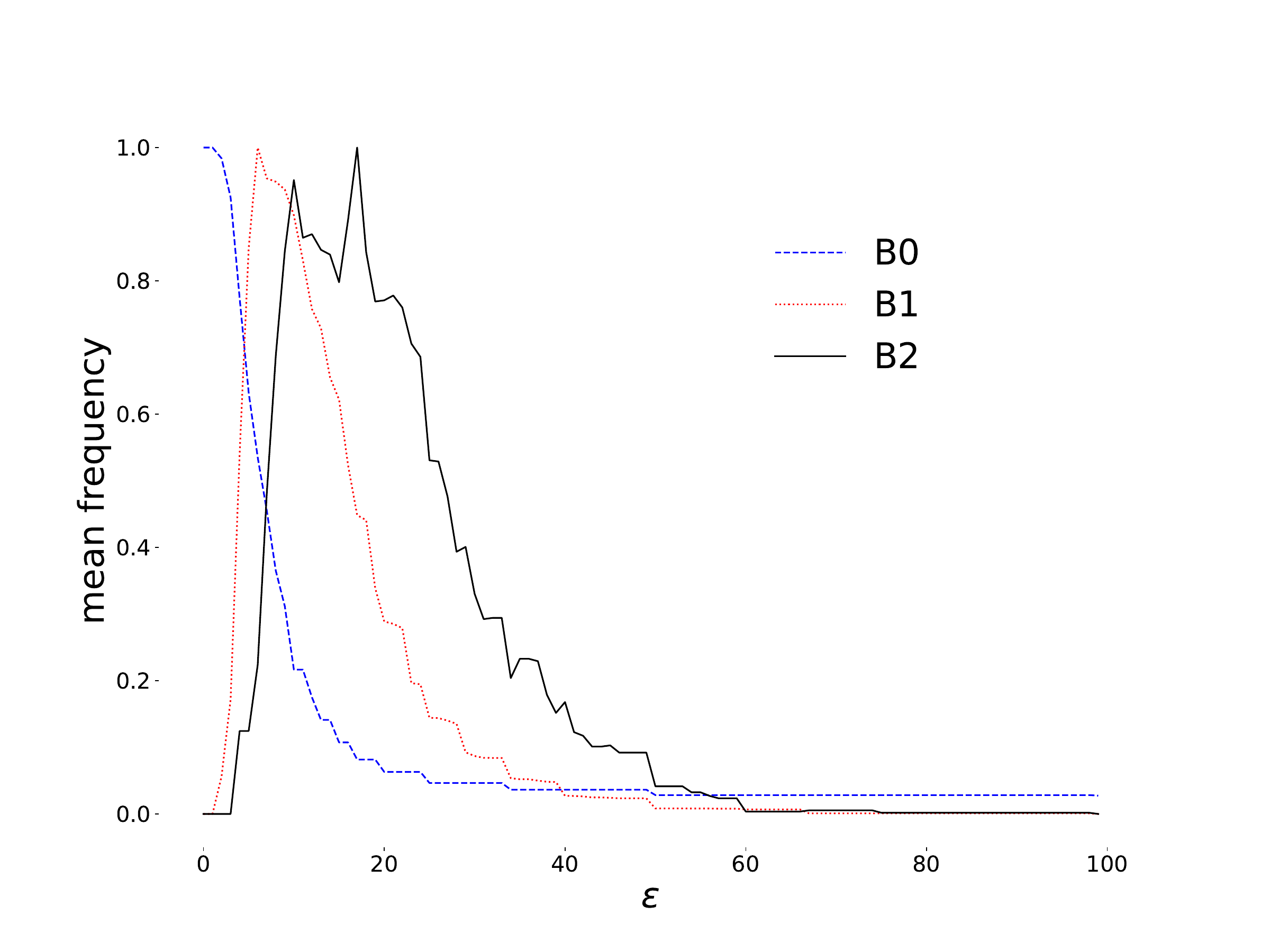}%
     \label{fig:ag}}
\hfil
\subfloat[REDDIT-MULTI-5K]{
     \includegraphics[width=0.3\textwidth]{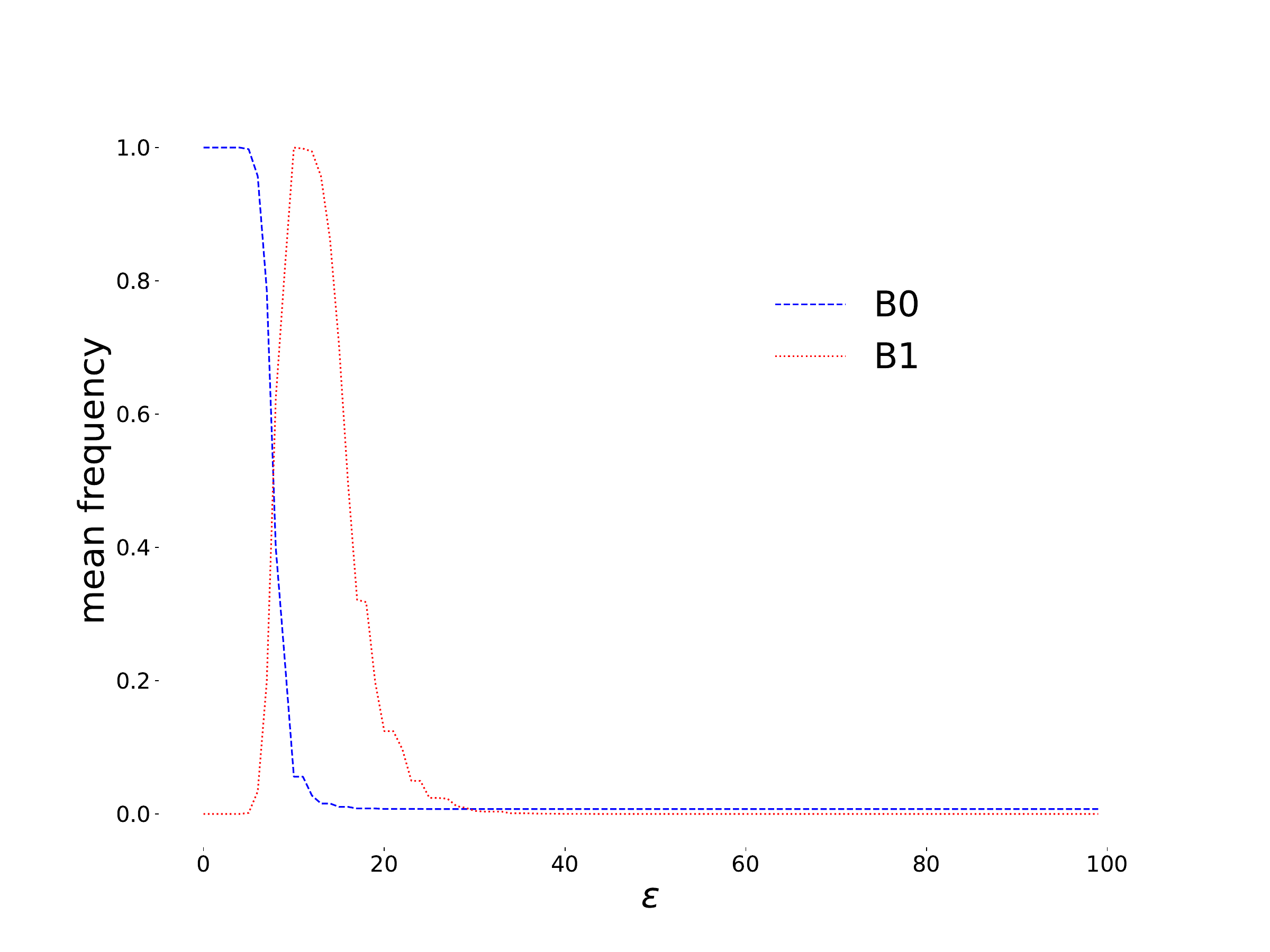}%
     \label{fig:ah}}
\hfil
\subfloat[REDDIT-MULTI-12K]{
     \includegraphics[width=0.3\textwidth]{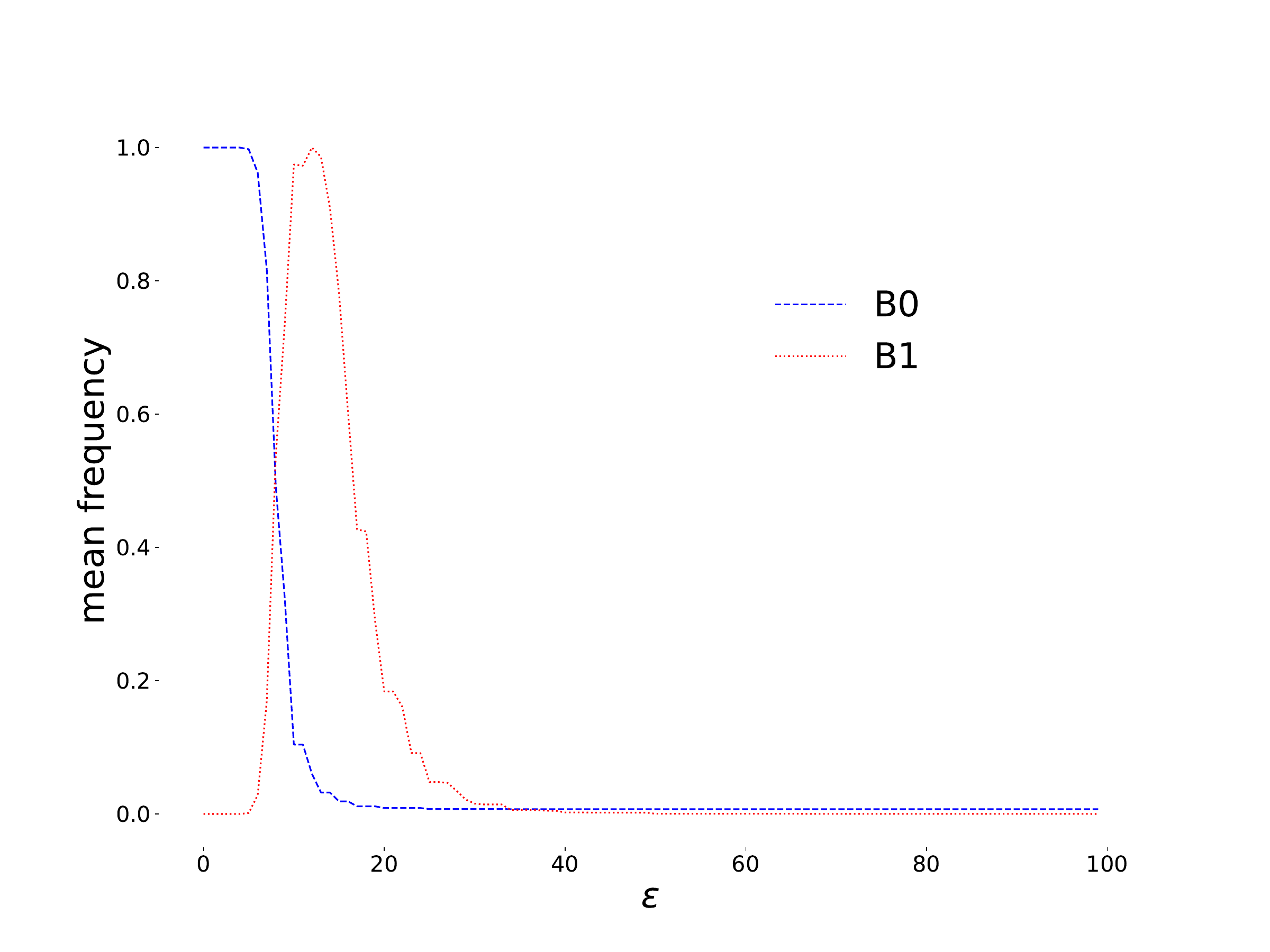}%
     \label{fig:ai}}
\caption{Normalized Betti Functions for Features of Dimensions 0, 1, and 2 Across All Datasets. Each Betti Function has Been Scaled by its Maximum Value Within the Respective Dataset.}
\label{fig:bettidistribution}
\end{figure*}

\section{Metrics used in evaluating our models}\label{sec:metrics}
\begin{itemize}
    \item Accuracy: We use accuracy in evaluating the performance of random forest in our classification task. It measures the proportion of correctly classified instances out of the total instances.
    \item Root Mean Squared Error (RMSE): We use this to evaluate the performance of the random forest regressor in our surrogate model formulation. It measures the magnitude of the errors predicted and actual values.
    \item Mean Absolute Error (MAE): Just like RMSE, we use MAE to evaluate the random forest regressor. It measures the average absolute difference between predicted and actual values.
    \item R-squared (R2) Score: We use R-squared as a statistical measure to represent the proportion of the variance in the dependent variable that is explained by the independent variables in our surrogate model. From our results, we see that higher values 0.816 and 0.7 for B1 and B2 respectively indicate that our surrogate model is a good fit for the data used.
\end{itemize}

\section{Additional Results in Graph Classification}\label{sec:additionalresults}
\noindent This section contains additional results in graph classification which includes edge deletion scenario when Vietoris-Rips-Betti with resistance distance is the filtering function, useful filtration ranges for B0, resistance distance Shap plot and decision tree for B2.

 \begin{figure}[h!]
     \centering
     \includegraphics[width=0.3\textwidth]{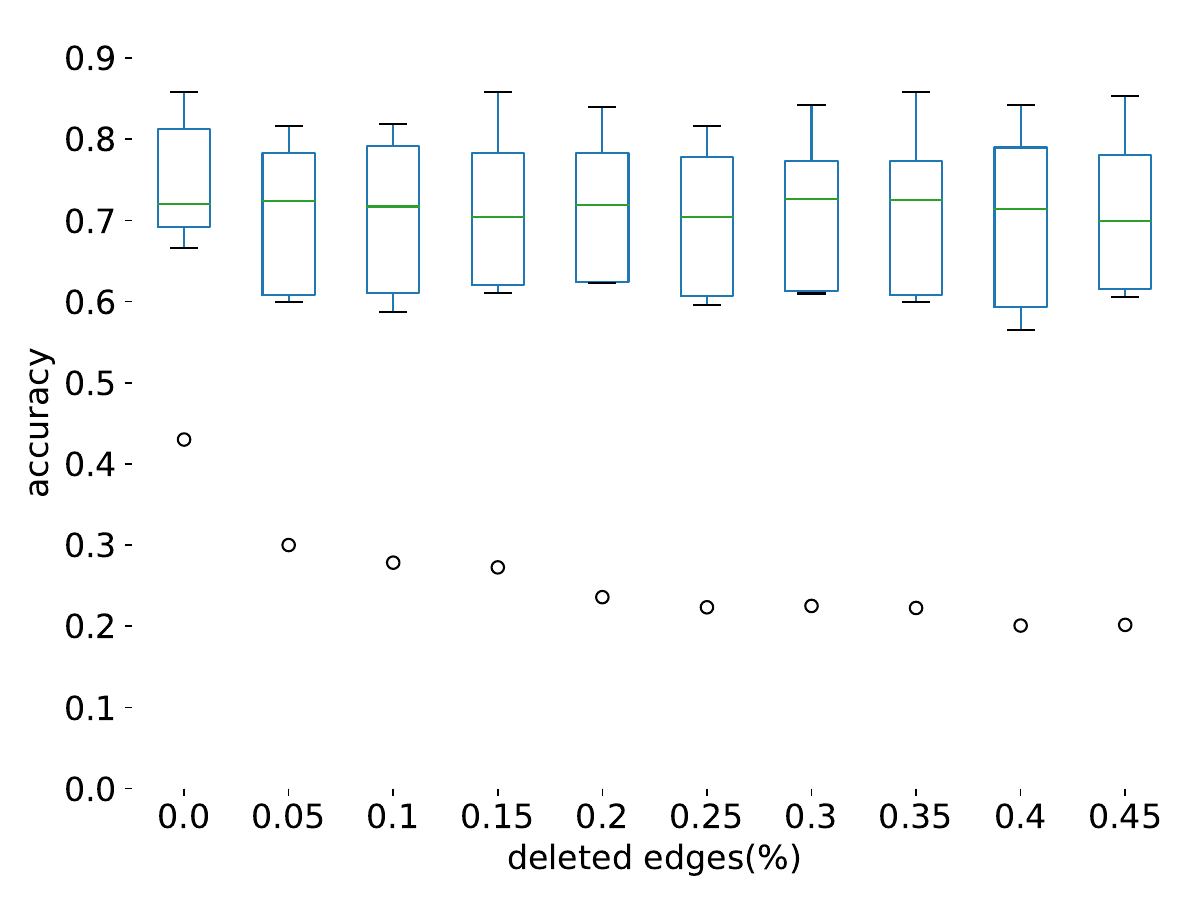}
    \caption{Average Accuracy Values Over the Nine Graph Datasets for Different Edge Deletion Scenarios. The Filtering Function used is Vietoris-Rips-Betti with Resistance Distance.}
     \label{fig:graphresistencerobustness}
 \end{figure}

\begin{figure}[h!]
\centering
\includegraphics[width=0.3\textwidth]{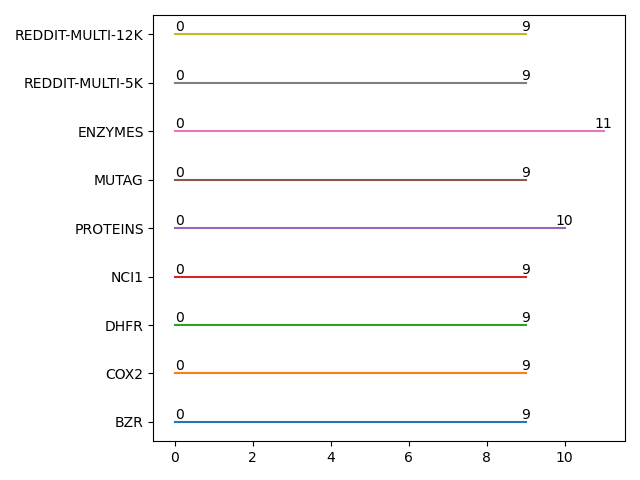}
\caption{Useful Filtration Ranges for B0. The x-axis Values are the Filtration Thresholds (in $[0,100]$) Defined Over Shortest Path Distance Between Node Pairs, and a Longer Range Between the Minimum and Maximum Values Indicates Greater Variability and Importance of These Thresholds in Influencing the Model's Predictions.}
\label{fig:b0}
\end{figure}

\begin{figure}[!t]
\centering
\subfloat[0-dimensional topological features]{
     \includegraphics[width=0.3\textwidth]{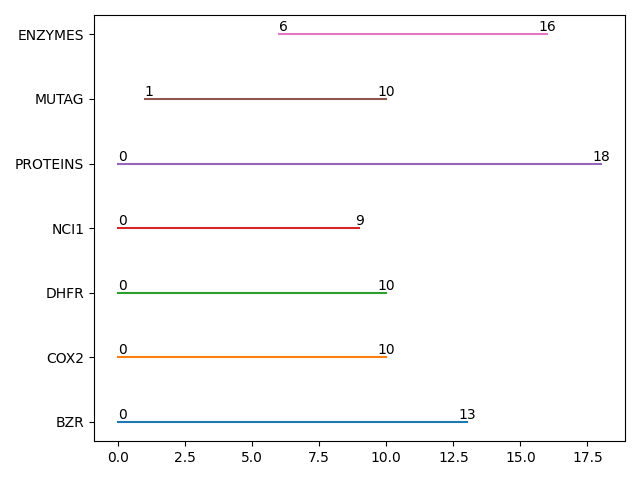}%
     \label{fig:restb0}}
\hfil
\subfloat[1-dimensional topological features]{
     \includegraphics[width=0.3\textwidth]{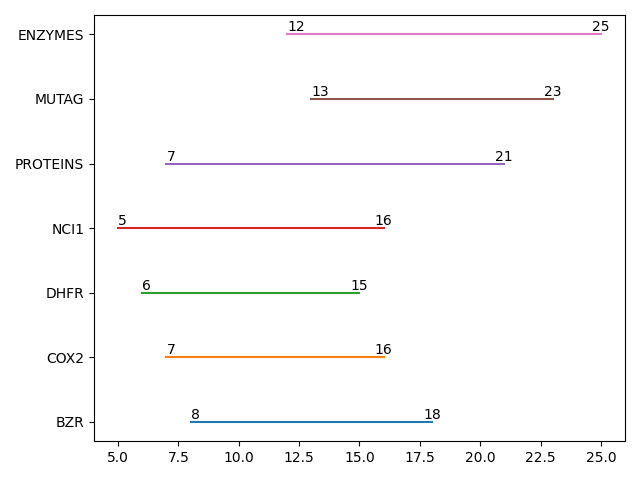}%
     \label{fig:restb1}}
\caption{The Range Between the Minimum and Maximum Values of the 10 Most Important Filtration Thresholds Identified Through the Shapley Analysis. The x-axis Represents the Filtration Thresholds (in $[0,100]$) Defined Over \textbf{Resistance Distance} Between Node Pairs, and a Longer Range Between the Minimum and Maximum Values Indicates Greater Variability and Importance of These Thresholds in Influencing the Model's Predictions.}
\label{fig:restimpstep}
\end{figure}

\begin{figure}[!t]
\centering
\includegraphics[width=\linewidth]{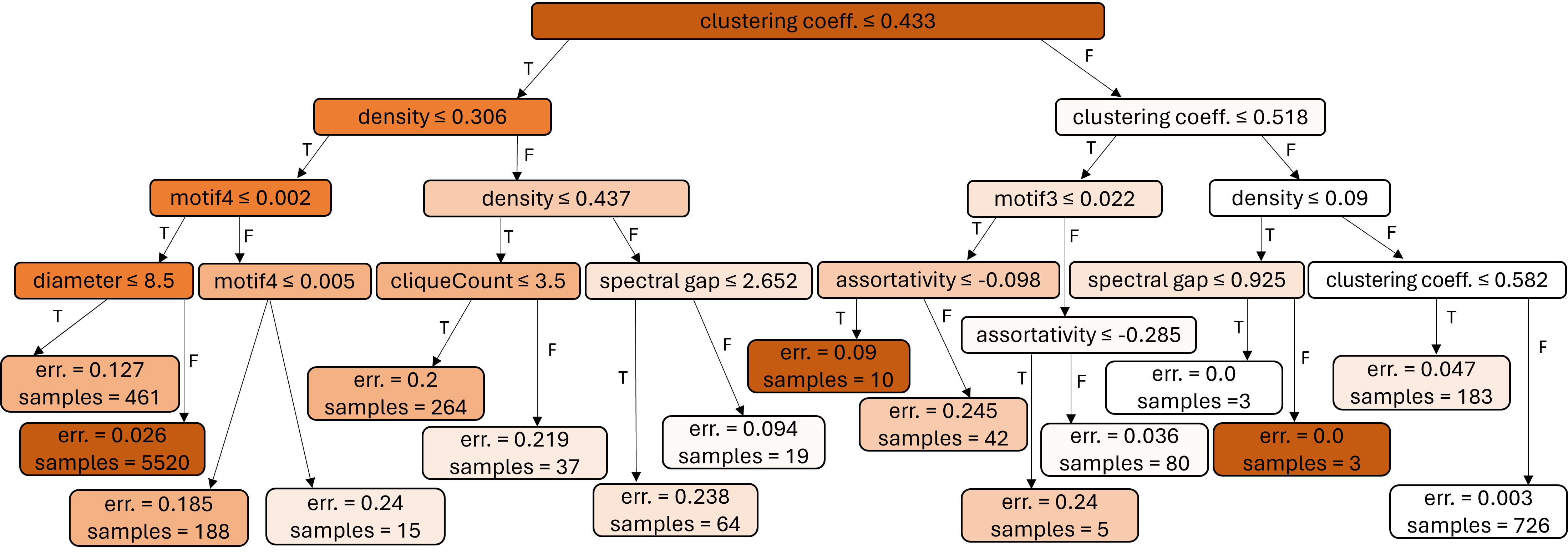}
\caption{Decision Tree Showing Features Split Using Decision Tree Regressor when the Target is Betti 2.}
\label{fig:decisiontreeb2}
\end{figure}

\section{Additional Robustness Experiment}
\noindent We performed additional experiment for the filtration method by removing 1\% to 5\% of the edges respectively and using the random classifier for the classification task. The table below shows the results obtained for NCI1 dataset:

\begin{table}[!t]
\caption{Table Showing Results for NCI1 when edges are removed up to 5\%.}
\label{tab:addrobustness}
\resizebox{9cm}{!}{
\centering
\begin{tabular}{lllll}
\toprule
 Edge removal(\%) & spd & WL & Dummy & GS  \\
\midrule
 0 & 67.54 $\pm$ 1.47 & 71.80 $\pm$ 0.32(3) & 49.25  $\pm$ 0.94 & 69.04  $\pm$ 1.30\\
 0.01 & 66.38  $\pm$ 1.07 & 72.48  $\pm$0.58(3) & 48.88  $\pm$ 0.81 & 65.18  $\pm$ 1.12\\
 0.02 & 64.66  $\pm$ 2.12 & 67.53  $\pm$0.42(2) & 48.76  $\pm$ 0.87 & 66.48  $\pm$ 1.40\\
 0.03 & 65.16  $\pm$ 1.53 & 68.55  $\pm$0.41(3) & 48.76  $\pm$ 0.93 & 66.27  $\pm$ 1.35\\
 0.04 & 65.16  $\pm$ 2.50 & 67.24  $\pm$0.58(2) & 48.75  $\pm$ 1.04 & 65.54  $\pm$ 1.47\\
 0.05 & 64.28  $\pm$ 1.71 & 67.04  $\pm$0.45(2) & 49.22  $\pm$ 0.54 & 65.37  $\pm$ 1.71\\
\bottomrule
\end{tabular}}
\end{table}

\section{TDA Mapper analysis}\label{sec:mapper}
 \noindent We employ the highly customizable TDA tool\textemdash Mapper~\cite{singhtda} to analyze and cluster graphs of the graph datasets. TDA Mapper complements traditional clustering, and projection pursuit approaches with a systematic insight into data geometry and topology. It uncovers hidden data patterns that are otherwise inaccessible with conventional data analytic techniques.

The key idea behind TDA Mapper is as follows: Let $T$ be a total number of observed graphs and $\{\vec{e}_t\}_{t=1}^T\in \mathcal{R}^{D^\prime}$ be a data cloud of graph features. For our dataset, $D^\prime=9$ (i.e., the nine graph features of each graph in a dataset). We employ the t-distributed stochastic neighbor embedding (t-SNE)~\cite{van2008visualizing} as a lens to reduce the data into a two-dimensional space. The t-SNE converts similarities between data points to joint probabilities and minimize the Kullback-Leibler divergence between the joint probabilities of the low-dimensional embedding and the high-dimensional data. Next, we select a function $\xi: \{\vec{e}_t\}_{t=1}^T \to \mathbb{R}$ that filters data in one of the two dimensions.

Let $I$ be the range of $\xi$, that is, $I = [m, M] \in \mathbb{R}$, where $m = \min \xi(\vec{e}_t)$ and $M = \max \xi(\vec{e}_t)$ in the dimension $d^\prime$. We place data into overlapping bins by dividing the range $I$ into a set $S$ of smaller overlapping intervals of uniform length and let $u_j= \{t : \xi(\vec{e}_t) \in I_j\}$ be graphs corresponding to features in the interval $I_j \in S$. For each $u_j$ we perform a k-means clustering to form clusters $\{t_{jk}\}$.

We analyze the empirical distribution of edge lengths where each cluster is merged to find the number of clusters. The merging criteria are based on the rationale that internal distances (i.e., within a cluster) are expected to be lower than external distances (i.e., in-between clusters), and distributions of internal and external distances are disjoint. Let $\{t_{jk}\}$ denote the $k$-th cluster of the $j$-th interval. We construct a cluster graph by transforming each cluster into a node and adding an edge between two nodes $k$ and $p$ if clusters  $\{t_{jk}\}$ and $\{t_{lp}\}$ contain overlapping data points, i.e., $\{t_{jk}\}\cap \{t_{lp}\}\neq \emptyset$. Formally, the graph is called a TDA Mapper graph or a topological network.

TDA Mapper produces a low dimensional representation of the underlying data structure in the form of \textquote{cluster tree} graph $\mathcal{CT}$ where each \textquote{cluster} is a branch of some single connected component rather than a disconnected component on its own as in conventional clustering analysis. This Mapper cluster network is shown  in figure~\ref{Mapper}.

\section{Related Work}\label{sec:relatedwork}
\noindent \input{relatedwork}

\vfill
\end{document}

%% file: introduction.tex
Over time, there has been an increasing trend of representing data as graphs~\cite{wu2020comprehensive}. Nevertheless, the irregularity and complexity of graph data have presented significant challenges for their integration into existing machine learning algorithms. To address these challenges, topological methods have been proposed, leading to successful applications in graph and node classification tasks~\cite{akcora2021bitcoinheist,chen2022time,chen2021tamp}.

Proponents of TDA argue that it excels in capturing intricate structures within data~\cite{chazal:inria-00292566}, such as loops in graphs. Moreover, TDA showcases robustness in handling noisy and high-dimensional datasets, ensuring reliable outcomes~\cite{carlsson2021topological}. Its interpretability enables the intuitive understanding of model behavior, while its stable nature instills confidence in the results. Being data agnostic, TDA easily adapts to diverse data types and allows seamless integration and fusion of multiple data sources, such as images, point clouds and graphs, promoting versatility and applicability across various domains.

We put these claims to test and conduct a comprehensive set of graph classification experiments to validate their assertions. Graph classification involves understanding the overall structure of a graph and making predictions based on that structure. A research area in topological data analysis, Persistent Homology~\cite{boissonnat2018geometric}, captures topological features of a graph across various scales, providing a global perspective on its shape and connectivity. This global understanding is particularly valuable for tasks where the arrangement of nodes and edges matters more than individual node or edge properties. For these reasons, we evaluate the effectiveness of persistent homology in the realm of graph classification, rather than edge or node classification. 

Our results substantiate that TDA indeed exhibits robustness against outliers and demonstrates interpretable behavior, aligning with the proponents' arguments. However, we also observe that TDA does not significantly enhance the predictive power of existing methods. Based on these findings, we propose the utilization of a surrogate model that incorporates TDA into the graph machine learning pipeline. This surrogate model aims to leverage the valuable insights offered by TDA while addressing the limitations identified in our evaluation. Through TDA, we also demonstrate how graph datasets are related to each other, revealing underlying connections and similarities that can enrich our understanding of the data landscape.

\begin{figure}[!t]
\centering
\includegraphics[width=3.5in]{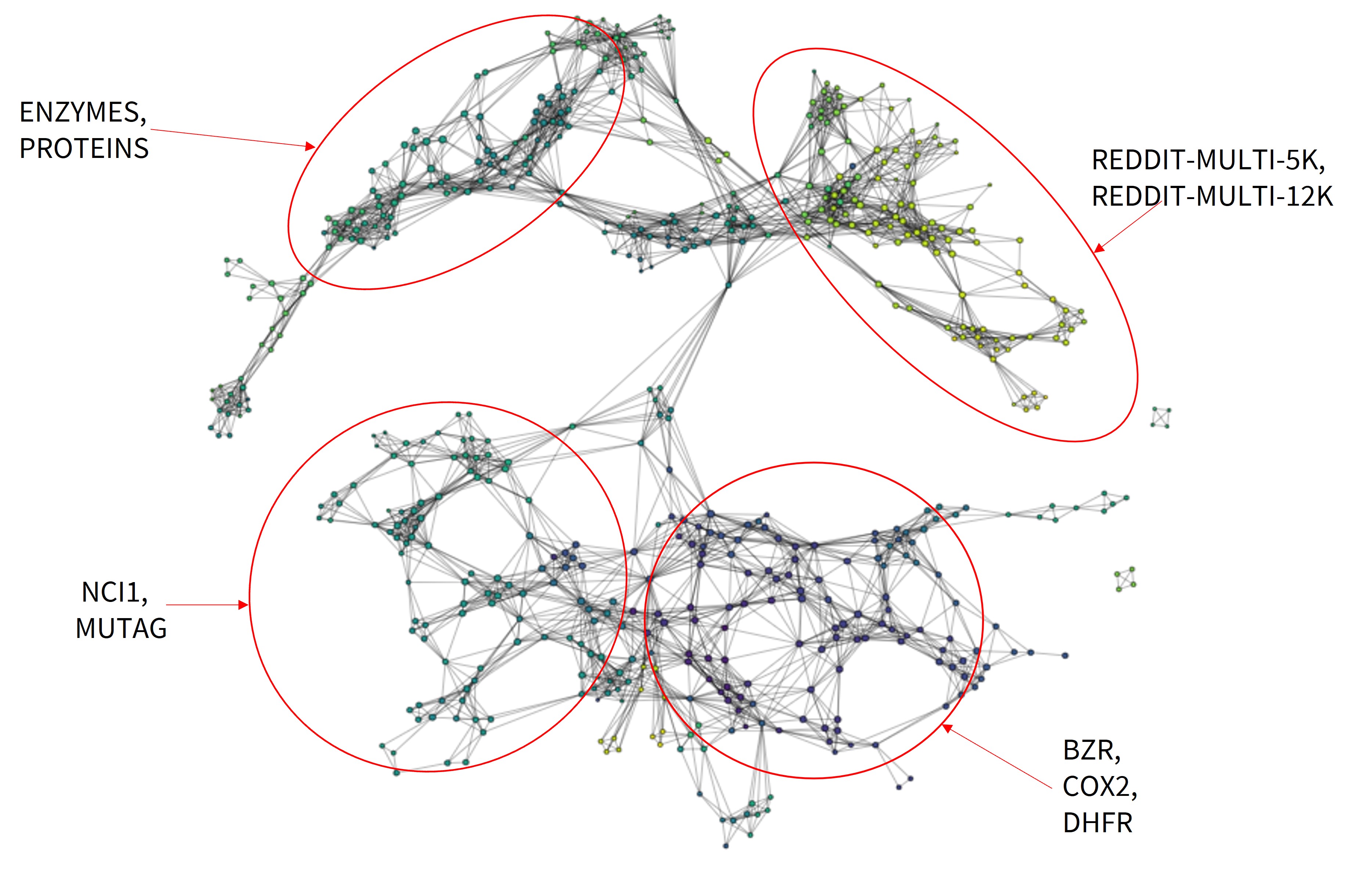}
\caption{A Topology-Based View on Datasets. The Center Contains $\texttt{NCI1}$ Graphs.}
\label{Mapper}
\end{figure}

Our contributions are as follows: 
\begin{itemize}
    \item Rigorous Testing: We rigorously test TDA claims regarding predictive power, complementarity, interpretability, and robustness in graph classification, providing practical insights into TDA's real-world effectiveness.
    \item Strengths and Weaknesses: Through our experiments, we systematically identify the strengths and weaknesses of TDA in graph classification, providing researchers with valuable insights for informed decision-making.
    \item Enhanced Understanding: We pinpoint a crucial factor in the TDA process, the localization of topological information within a concise filtration range. This discovery holds the potential to facilitate the scalability of TDA to larger graphs.
    \item Surrogate Model: Our proposed surrogate model strategically integrates TDA to improve the interpretability of graph classification, allowing visualization of valuable features and addressing limitations while capitalizing on TDA's strengths.
\end{itemize}

%% file: background.tex
In this section, we present the methodology for the graph classification problem, as outlined in~\cite{hensel2021survey}. Section~\ref{sec:tutorial} offers an independent and quite accessible tutorial that covers essential topological concepts required to follow this article. Due to space constraints, we have included the related work in Appendix~\ref{sec:relatedwork}.  

\noindent\textbf{Research Problem [Graph Classification]}. Assume a collection of graphs $G = \{G_1, G_2, ..., G_N\}$ where each $G_i = (V_i, E_i)$ is described as having vertices $V_i$ and edges $E_i$.  Graph classification aims to learn a function $f: G_i \rightarrow L$, where $L$ is the set of graph labels. We examine topological methods, and subsequently present Weisfeiler-Leman (WL) graph kernels~\cite{shervashidze2011weisfeiler} and  graph feature-based classifiers as benchmarks, for the purpose of contrasting the obtained results. 

\noindent\textbf{Kernels and GNNs.} Our choice of using the Weisfeiler-Leman kernel as a benchmarking method is due to an important result in graph classification, which states that popular message passing graph neural networks cannot distinguish between graphs that are indistinguishable by the 1-WL test~\cite{morris2019weisfeiler}. For this reason, we use the WL kernel as a baseline of what message passing GNNs can achieve and compare its performance to the TDA approaches. The state-of-the-art results for each dataset mainly come from GNN methods. While we do present these results, our primary focus is on comparing TDA methods with kernel and feature-based classifiers. Comparing graph neural networks directly with more cost-effective TDA-based methods is unfair due to the significant computational expenses of GNNs.

%% file: topology.tex
Topology is concerned with analyzing the shape of data, and likewise, topological graph machine learning focuses on shapes that can be embedded within graphs. Persistent Homology research in TDA studies shapes of increasing dimensions, such as connected components, loops, and voids (i.e., regions that are surrounded by higher-dimensional structures). For instance, a basic graph loop that connect vertices such as $v_1\rightarrow v_2 \rightarrow v_3 \rightarrow v_4 \rightarrow v_1$ is a one-dimensional hole in topology. Higher topological features such as voids, holes, are not as straightforward to explain in graph terminology as the loop example.  

\begin{figure*}[!t]
    \centering
    \includegraphics[width=\linewidth]{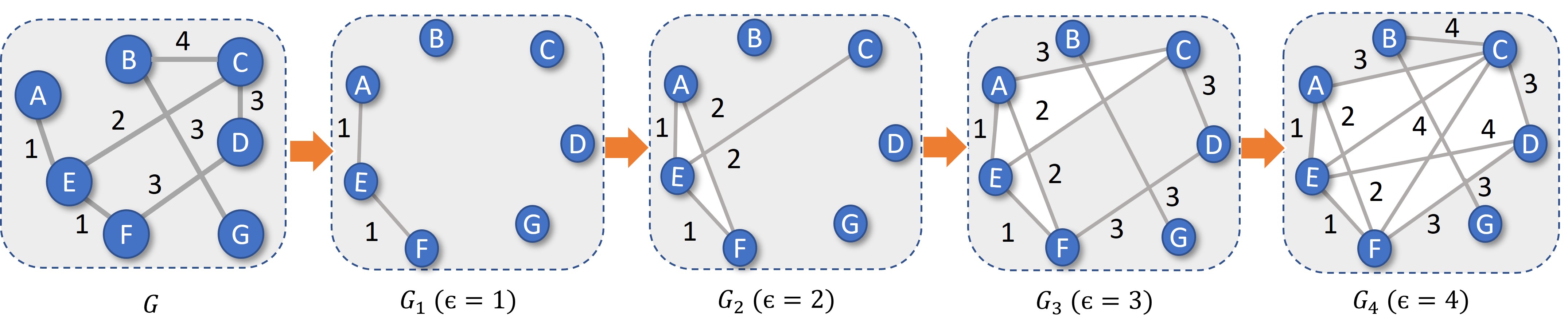}
    \caption{Vietoris-Rips Filtration. A Weighted Graph $\G$ is Embedded Into a Metric Space where the Distance Between Two Nodes is Defined as the Shortest Path Between Them. $\G_1,\ldots,\G_4$ are the First Four Complexes of the Resulting Vietoris-Rips Filtration, where the Nodes are 0-Simplices, the Edges are 1-Simplices and (Filled) Triangles are 2-simplices. At $\epsilon=3$, a One-Dimensional Hole is Formed by Edges $C-E$, $C-D$, $D-F$, and $E-F$, but gets Filled at $\epsilon=4$ when Triangles (2-Simplices) $\triangle{CDE}$ and $\triangle{DEF}$ are added to $G_4$.}
    \label{fig:filtration}
\end{figure*}

\noindent\textbf{Persistent Homology in GraphML~\cite{boissonnat2018geometric}.} Graph machine learning involves utilizing all the nodes and edges within a graph to develop a classification model. However, graphML may opt to focus on a particular set of nodes, such as those that are most central. Conversely, topology employs a \textit{filtration} technique to establish the nodes and edges of a graph, then identifies the shapes that arise during this filtration process as topological features. The filtration approach is akin to analyzing a temporal network over time, with topology providing a similar perspective through node or edge activations. 
In this section, we will discuss various techniques for graph filtering, along with several types of shapes and representations for conveying their evaluation along a filtration. In the rest of this section, we begin by formally presenting various topological concepts and their relevance to the task of graph classification.

Topology encompasses not only the shapes of objects, but also their birth and death. To extract topological information from graphs, we employ persistent homology, which involves constructing a \textit{filtration} of \textit{abstract simplicial complexes} on the graphs. 
 
\begin{definition}[Abstract Simplicial Complex] Assume a discrete set $X$. An abstract simplicial complex is a collection $\mathbb{C}$ of finite subsets of $X$ such that if $\sigma \in \mathbb{C},$ then $\tau \in \mathbb{C}$ for all $\tau \subseteq \sigma.$ If $|\sigma| = p+1,$ then $\sigma$ is a $p$-simplex (p $\geq$ 0). 
\end{definition}

A $p$-simplex in topology corresponds to a $p+1$ clique in graph theory. For instance, a 2-simplex is a (filled) triangle consisting of nodes A, B, and C, where edges $A-B$, $B-C$, and $C-A$ are present.  

\noindent\textbf{Simplicial Complex in GraphML.} Typically, graphs are created by connecting nodes using simple edges. However, there are situations where hyperedges, which connect multiple nodes at once, can also be employed in the graph construction process. In a simplicial complex, nodes (0-simplices) and edges (1-simplices) also exist, but there are additional higher-dimensional simplices as the building blocks. For example, a 2-simplex can be connected to a node through a simple edge.  Intuitively, a simplicial complex can be viewed as a higher dimensional generalization of graphs representing a structure consisting of nodes, edges, triangles, and their higher-order counterparts.

\subsection{Vietoris-Rips Complexes}
\label{sec:complex}
Our first choice in a simplicial complex is {\it Vietoris-Rips (VR)} because it is widely used due to its easy construction~\cite{CarlssonTopologyData2009}. 

 \begin{definition}[Vietoris-Rips Complex~\cite{vietoris1927hoheren}] Given a metric space $(X, d)$ and $\epsilon \geq 0,$ we denote the VR (abstract) simplicial complex on $k$ vertices $x_0,x_1,\dots,x_k$ by $\mathcal{M}_\epsilon$, where an edge existing between two vertices $x_i$ and $x_j$ is created if $d(x_i, x_j) \leq \epsilon$ for all $i \neq j$~\cite{kovacev2016using}. 
 \end{definition}

\subsection{Alpha Complexes}\label{sec:apxalphacomplex}
Certain complexes are based on a \textit{sparsification paradigm}, wherein a subset of the data points is used to create a persistence diagram that accurately represents the persistence diagram of the original data point cloud. This idea of sparsification is similar to graph coarsening techniques~\cite{liu2018graph}, where nodes and edges are removed from the graph to reduce computational costs. The Alpha complex~\cite{akkiraju1995alpha} is an example of such a complex, which was developed to address complexity concerns in low-dimensional settings. 

\begin{definition}[Alpha Complex] Suppose $\mathcal{K}$ is a finite set of points in a \textit{d}-dimensional space $\mathbb{R}^d$. The Voronoi cell of a point $k \in \mathcal{K}$ is the set of points for which $k$ is the closest, defined as \[V(k,\mathcal{K})  \coloneqq \{y \in \mathbb{R}^d \mid d(k,y)\leq d(k^{\prime},y), \, \text{for all} \, k^{\prime} \in K \}.\] In addition, we define a closed ball with center $k$ and radius $r$ as $\mathcal{B}_k(r)$. The intersection of each ball with the corresponding Voronoi cell is a Voronoi ball given as $\mathcal{R}_k(r) = \mathcal{B}_k(r) \cap V(k,\mathcal{K}).$ Then, the \textbf{Alpha Complex} of $\mathcal{K}$ is defined as the nerve of all the Voronoi balls, i.e. \[Alpha(r) \coloneqq \{\sigma \subseteq \mathcal{K} \colon \bigcap\limits_{k \in \mathcal{Q}} V(k,\mathcal{K}) \neq \emptyset \}.\]  
\end{definition}
\noindent\textbf{Alpha Complex in GraphML.}  Alpha complex must choose a subset of data points from an $n$-dimensional point cloud to compute persistent homology. The complex uses distances of data points in a cloud for computation, with its worst-case time as $O(k^2 \log k)$ for a dataset with $k$ points in 2D, and $O(k^3)$ for a dataset with $k$ points in 3D~\cite{otter2017roadmap}. Devising a pairwise node distance matrix for building Alpha complex in graphs leaves us with a distance matrix of $\left | \V\right |\times \left | \V\right |$ which would be too costly to use in the complex. As a result, we must employ a dimensionality reduction technique, such as t-SNE~\cite{van2008visualizing}, PCA~\cite{hotelling1933analysis}, or MDS~\cite{kruskal1964multidimensional}, to first reduce the $\mathbb{R}^{\left | \V\right |}$ dimensions into a more manageable dimension $\mathbb{R}^2$ or $\mathbb{R}^3$. The choice must consider the topological dimension that we are targeting; if we want to track 1D holes, the reduced form should be at least in $\mathbb{R}^2$. In general, we need data in $\mathbb{R}^d$ to track $(d-1)$-dimensional holes.  We demonstrate the effect of dimensionality reduction methods on the outcomes of the Alpha complex in Section~\ref{sec:experiments}.

\subsection{Filtration} 
\label{sec:filtration}
Having built simplicial complexes through VR or Alpha complex, our next task is to define a filtration over the complex. To this end, we fix a sequence of scale resolutions  $\epsilon_1<\epsilon_2<\ldots<\epsilon_n$ and form a chain of nested  VR or Alpha complexes called a \textit{finite filtration}. This process is illustrated in figure~\ref{fig:filtration} for the VR filtration built from a graph.

\begin{definition}[Filtration]
A filtration \( \mathcal{F} \) of a simplicial complex \textit{M} is defined as a collection of nested subcomplexes of \textit{M}, which moves from the empty set $\emptyset$ to \textit{M} in the following form:
\[\emptyset = \mathcal{F}_{0} \subseteq \mathcal{F}_{1} \subseteq \dots \subseteq \mathcal{F}_{r} \subseteq \dots \subseteq \mathcal{F}_{N} = M.\]
\end{definition}

\noindent
\(\mathcal{F}\) in a sense, approximates a dataset at different scales $n \in \{0,...,N\}$. 

The filtration process in topological data analysis often requires an appropriate distance metric between data points, which implies finding distances between node pairs in the context of graph analysis. In this paper, we define the distance in two ways: the shortest-path distance computed using Dijkstra's algorithm~\cite{johnson1973note} or the graph resistance-distance~\cite{babic2002resistance}. 

\noindent\textbf{Node Distance in GraphML.}
The shortest-path distance is a widely used metric in graph theory, which calculates the minimum number of edges needed to traverse from one node to another in a graph. 

Graph resistance-distance, also known as the effective resistance or commute time distance, is a concept in graph theory that measures the similarity or dissimilarity between nodes in a graph based on the effective resistance between them~\cite{babic2002resistance}. It provides a notion of distance or dissimilarity by considering the conductance of edges in the graph. In electrical circuit theory, the resistance between two points in an electrical network is a measure of the difficulty for electrical current to flow between those points. Analogously, the resistance-distance between two nodes in a graph is a measure of the difficulty for a ``random walk" to move between those nodes. The resistance-distance between two nodes $u$ and $v$ in a graph $\G$ is defined as the effective resistance between the nodes when an ideal unit current is injected at $u$ and absorbed at $v$, while all other nodes in the graph are grounded (set to zero potential). Mathematically, it is computed as the voltage difference between $u$ and $v$ when a unit current is injected at $u$. The resistance-distance takes into account the conductance of edges and the connectivity of the graph. Shorter resistance-distances typically indicate nodes that are more structurally similar or more easily reachable in the graph.

Once it is computed, the distance matrix serves as a fundamental data structure on which the VR and Alpha complexes are constructed and the filtration is obtained. Filtration tracks the appearance of shapes along the distance threshold $\epsilon$, and stores the information gathered in a \textit{persistence diagrams} (PD), which consist of a multiset of points supported on the open half-plane ${(\alpha_b, \alpha_d) \in \mathbb{R}^2 :\alpha_b < \alpha_d } \subset \mathbb{R}^2$~\cite{de2022ripsnet}. Here, $\alpha_b$ and $\alpha_d$ respectively represents the \textit{birth} and \textit{death} of new topological features, such as connected components, loops, cavities, and others, that arise and vanish during the filtration process. By keeping track of the birth and death times of these features (figure ~\ref{fig:mutagPD}), we can understand their evolution over different scales and extract valuable insights about the underlying data structure.

\noindent\textbf{Persistence Diagram in GraphML.} Research articles commonly showcase persistence diagrams that exhibit zero-dimensional and one-dimensional shapes, along with their birth and death times (i.e., $\epsilon$ values). For instance, in Appendix figure~\ref{fig:mutagPD}, birth and death times are presented for zero-dimensional (circles) and one-dimensional (stars) holes. Zero-dimensional holes correspond to connected components of graphs where no edges are activated yet, while one-dimensional holes are graph loops that may involve three or more nodes which typically emerge later and vanish when the distance threshold $\epsilon$ exceeds the path distance between the farthest nodes in the loop. As more edges come into the filtration, a connected graph will only retain one zero-dimensional hole, which persists.  
If a hole emerges and disappears shortly after, it will be located closer to the diagonal line while a hole that persists has a longer lifespan.

\noindent\textbf{Complexity of Persistent Homology:\label{sec:complexity}} The computational complexity of the $k^{th}$ persistence diagram ($PD_k$) is $\mathcal{O}(n^3)$ where $n$ is the number of $k$-simplices~\cite{otter2017roadmap}.~\cite{mischaikow2013morse} achieves $\mathcal{O}(m^2\times n\log{n})$ where $m$ is the number of critical $k$-simplices. With the additional time to find the critical $k$-simplices in each filtration step, the computational complexity  $\mathcal{O}(m^2\times n\log{n})$ is still not scalable for very large networks.  In general, persistent homology is subject to the same computational limitations as graph neural networks, and is best suited for graphs containing fewer than 1,000 nodes.

\subsection{Vectorizations of Persistence Diagrams}
PDs are useful shape descriptors of data~\cite{hensel2021survey}. However, due to their non-vector form, we cannot use them directly as input in ML pipelines. A common remedy to address this limitation is to construct a summary function from a given PD and vectorize it using a sequence of scale (or distance) values. A typical way of extracting ML-usable information from a PD is to track topological shapes as a function of the distance threshold $\epsilon$. Specifically, the Betti numbers~\footnote{In persistent homology, the Betti number is named after Enrico Betti, an Italian mathematician who lived in the 19th century. Betti was known for his work in algebraic topology, which is the branch of mathematics that studies topological spaces using algebraic methods.}, denoted as $\beta_p$, are a numerical invariant of a topological space that describe the number of p-dimensional holes in the space. Consider a dataset where two 1-dimensional holes (i.e., a graph loop) are born at $\epsilon=0.5$ and they die at $\epsilon=1.5$. We say that for any $t,$ $0.5 \leq t < 1.5$, $\beta_1(t)=2$, i.e. two 1-dimensional holes exist at $\epsilon=t$. This implies that we can create a function of Betti values over the filtration threshold $\epsilon$. We use a vectorized Betti function which is one of the simplest functional summaries of PDs~\cite{edelsbrunner2022computational}. There exist two other commonly used vectorization methods,  persistent landscapes, persistent silhouette ~\cite{otter2017roadmap, hensel2021survey} to summarize the information in a PD. We define the mentioned summary functions as follows:

\begin{definition}[Betti Function]\label{Betti}
	Let $D=\{(b_i,d_i)\}_{i=1}^n$ be a persistence diagram (of homological dimension $p$). The Betti function (associated with $D$ and $p$) is defined as 
	\begin{equation}\label{eqn:betti_function}
	\beta_p(t)=\sum_{i=1}^nw(b_i,d_i)\chi_{[b_i,d_i)}(t),    
	\end{equation}
	where $w:\{(x,y)\in\mathbb{R}^2:x\leq y\}\rightarrow \mathbb{R}$ is a weight function and $\chi_{[b,d)}(t)=1$ if $t\in[b,d)$ and 0 otherwise. 
\end{definition}

The function $w$ that assign weights to simplices based on their persistence ($d-b$) is often non-decreasing, and we set it to $w\equiv 1$ in our experiments. The resulting function $\beta_p(t)$ is referred to as the $p$-th Betti number which is the count of $p$-dimensional holes in the simplicial complex at a given scale $t$ within a filtration, that are still present and \textquote{born} at or before $t$.

\begin{definition}[Persistence Landscape (PL)]\cite{bubenik2015statistical}\label{PL}
The $k$-th order landscape function of a persistence diagram $D=\{(b_i,d_i)\}_{i=1}^n$ (of homological dimension $p$) is defined as 
$$
\lambda_p(k,t) \coloneqq \hbox{kmax}_{1\leq i\leq n}f_{(b_i,d_i)}(t),
$$ 
where $f_{(b,d)}(t) = max\{0,min\{t-b,d-t\}\}$ and $\hbox{kmax}$ denotes the $k$-th largest element of the set. 
\end{definition}  

Persistence Silhouette (PS) is a variation of a PL which takes the form of a weighted average of the \textit{tent} \footnote{Tent functions are a type of piecewise linear function that resemble a triangular tent shape. They are defined by dividing the domain into equal intervals and then linearly increasing and decreasing the value of the function within each interval, reaching a maximum value at the midpoint and then decreasing back to zero. Tent functions are commonly used in signal processing, numerical analysis, and computer graphics.} functions $f_{(b,d)}(t)$ given in Definition~\ref{PL}.
\begin{definition}[Persistence Silhouette (PS)]\cite{chazal2014stochastic}  The $k$-th power silhouette function of a persistence diagram  $D=\{(b_i,d_i)\}_{i=1}^n$ (of homological dimension $p$) is defined by
$$
\phi_p^{(k)}(t) = \dfrac{\sum_{i=1}^n |d_i - b_i|^k f_{(b_i,d_i)}(t)}{\sum_{i=1}^n |d_i - b_i|^k}.
$$
\end{definition}

A summary function is vectorized by evaluating it at each point of an increasing sequence of scale values $\{t_1,t_2,…,t_d\}$ to produce a vector in $\mathbb{R}^d$.

%% file: tables/accuracytable.tex
\begin{table*}[h!]
\caption{Graph Classification Accuracy Results Obtained Using Vietoris-Rips (VR), Alpha Complex (AC), Weisfeiler-Lehman Kernel (WL), and Graph Features (Baseline). The Vectorizations Used are Betti functions (B), Persistence Landscape (L), and Persistence Silhouette (S). Bold Represents Best Results while Second-Best Results are \underline{underlined} Among the TDA, Kernel (h is the number of iterations) and Baseline Methods (i.e., SOTA GNN Results are Excluded). \label{tab:firstexperiment}}
\centering

\resizebox{18cm}{!}{
\begin{tabular}{c l c c c c c c c c c}
\toprule
&Model&BZR&COX2&DHFR&ENZYMES&MUTAG&NCI1&PROTEINS&RED-5K&RED-12K\\
\midrule
 
\parbox[t]{2mm}{\multirow{4}{*}{\rotatebox[origin=c]{90}{TDA}}}&VR-B &\textbf{86.30 $\pm$ 2.82}&$78.94 \pm 2.74$&$71.97\pm 4.22$&$\underline{43.00\pm 3.12}$&$85.79\pm 7.57$&$66.59\pm 1.34$&$72.24 \pm 2.88$&\underline{52.54 $\pm$ 1.40} &\textcolor{red}{OOR}\\

&VR-L &$79.38\pm 3.73$&$78.40\pm 4.32$&$60.86\pm 3.40$&$23.67\pm 3.36$&$74.47\pm 4.48$&$57.69\pm 1.25$&$61.39\pm 2.65$&$28.58\pm 1.69$&$22.75\pm 0.44$\\

&VR-S &$81.85\pm 3.08$&$79.04\pm 3.25$&$61.25\pm 2.02$&$27.83\pm 4.09$&$78.16\pm 4.12$&$62.26\pm 1.27$&$64.04\pm 2.65$&$39.53\pm 1.13$&$26.28 \pm 0.92$\\
  
&AC-B &$77.90\pm 3.97$&$78.09 \pm 2.80$&$73.62 \pm 2.85$&$28.25\pm 3.59$&$82.63 \pm 3.55$&$64.17 \pm 1.19$&$73.23 \pm 2.95$&$47.89\pm 1.09$&$39.20\pm 0.91$\\

&AC-L &$81.61\pm 3.21$&$76.38\pm 4.07$&{73.82 $\pm$ 3.78}&$22.33\pm 3.58$&$73.68\pm 5.69$&$62.19\pm 1.35$&$64.84\pm 4.58$&$42.18\pm 1.34$&$30.46\pm 0.90$\\

&AC-S &$80.00\pm 3.38$&$77.02\pm 4.49$&$\underline{74.61\pm 3.63}$&$21.92\pm 4.08$&$73.95\pm 6.50$&$62.23\pm 2.52$&$64.57\pm 2.93$&$42.73\pm 1.82$&$27.97 \pm 1.04$\\
 
\midrule
 & WL(h) &$83.20\pm 1.19(2)$&$\textbf{84.47 $\pm$ 1.03(2)}$&$\textbf{80.07 $\pm$ 1.12(3)}$&$\textbf{51.67 $\pm$ 1.30(3)}$&$\textbf{88.16$\pm$ 2.24(2)}$&$\textbf{79.49 $\pm$ 0.38(3)}$&$\textbf{76.28 $\pm$ 0.39(2)}$&$46.94 \pm 0.28(3)$&\underline{37.53 $\pm$ 0.50(2)}\\ 
 
 \midrule 
 &Baseline & $\underline{84.94\pm 2.72}$ &$\underline{79.57 \pm 3.36}$&${74.01 \pm 3.99}$&$40.50 \pm 2.87$&$\underline{87.89 \pm 4.68}$&$\underline{69.96 \pm 0.85}$&$\underline{73.32 \pm 2.92}$&$\textbf{54.29 $\pm$ 0.79}$&$\textbf{45.66 $\pm$ 0.89}$\\

 \midrule 
 &SOTA &$90.9 \pm 3.2$ \cite{chen2022redundancy} &$87.6 \pm 4.0$ \cite{xu2018powerful} & $84.5 \pm 4.6$ \cite{chen2022redundancy} & $75.3 \pm 5.0$ \cite{chen2022redundancy} &$93.3 \pm 6.0$ \cite{chen2022redundancy}& $83.6 \pm 1.6$ \cite{chen2022redundancy} &$77.5 \pm 3.4$ \cite{chen2022redundancy}& $57.5 \pm 1.5$\cite{xu2018powerful} & $44.5$\cite{hofer2017deep} \\
 
\bottomrule
\end{tabular}
}
\end{table*}

%% file: tables/detail01table.tex
\begin{table*}[bt!]
\caption{Graph Classification Accuracy Results Obtained Using Vietoris-Rips (VR), Alpha Complex (AC) Using Either 0- or 1-dimensional Homology (H). We use Betti Function Vectorizations in the VR (VR-B) and AC (AC-B) Models. \label{tab:Betti0and1experiment}}
\centering
\resizebox{18cm}{!}{
\begin{tabular}{c|c c c c c c c c c c }
\toprule
Dimension&Model&BZR&COX2&DHFR&ENZYMES&MUTAG&NCI1&PROTEINS&RED-5K&RED-12K\\
\midrule
\parbox[t]{2mm}{\multirow{4}{*}{\rotatebox[origin=c]{90}{0-Dim}}}&VR-B &$80.25 \pm 4.83$&$77.13 \pm 3.48$&$67.57\pm 3.64$&${39.17 \pm 2.52}$&$83.68\pm 5.66$&$65.05\pm 1.82$&$72.20\pm 2.83$&$41.42\pm 1.53$&\textcolor{red}{OOR}\\

&AC-B &$81.60 \pm 3.26$&$78.19\pm 3.66$&$69.61\pm 4.41$&$29.92\pm 2.50$&$80.00 \pm 6.23$&$64.11\pm 1.30$&$\mathbf{74.48 \pm 2.64}$&$46.05\pm 1.36$&$38.64\pm 0.81$\\

&Base + VR-B &$82.59 \pm 3.92$&${78.30 \pm 5.66}$&$70.72\pm 3.30$&$38.92\pm 4.21$&$84.21\pm 4.11$&$67.87 \pm 1.43$&$73.23 \pm 2.48$&${53.41 \pm 1.33}$ &\textcolor{red}{OOR}\\

&Base + AC-B &$\mathbf{86.05 \pm 1.31}$&$76.92\pm 4.11$&$\mathbf{74.87 \pm 4.25}$&$\mathbf{41.25 \pm 5.22}$&${87.11 \pm 6.13}$&${68.56 \pm 1.32}$&$73.64\pm 1.87$&$52.17\pm 1.76$&$\underline{44.40 \pm 1.05}$\\

\midrule

\parbox[t]{2mm}{\multirow{4}{*}{\rotatebox[origin=c]{90}{1-Dim.}}}&VR-B &$83.33\pm 3.87$&$\underline{79.15 \pm 3.22}$&$67.90\pm 4.03$&$31.67 \pm 3.66$&$\mathbf{89.47 \pm 5.41}$&$63.20 \pm 1.15$&$61.08 \pm 1.92$&$45.98\pm 1.30$ &\textcolor{red}{OOR}\\

&AC-B &$80.25\pm 3.24$&$74.47\pm 3.33$&$68.62\pm 2.72$&$17.33\pm 3.47$&$73.42 \pm 5.47$&$60.40 \pm 1.05$&$65.11\pm 3.86$&$43.74\pm 0.86$&$30.33\pm 0.65$\\

&Base + VR-B &$84.57 \pm 5.01$&$77.98\pm 2.93$&$\underline{74.34 \pm 3.10}$&$38.42 \pm 2.79$&$87.37 \pm 2.72$&$\underline{68.77 \pm 1.80}$&${73.96 \pm 3.05}$&$52.96\pm 0.68$ &\textcolor{red}{OOR}\\

&Base + AC-B &$\underline{85.93 \pm 2.92}$&$77.02\pm 4.20$&$74.15 \pm 4.43$&$38.33\pm 2.12$&$86.32 \pm 5.52$&$66.84\pm 1.89$&$\underline{74.23 \pm 1.86}$&$\underline{53.92 \pm 1.60}$&${44.02 \pm 0.99}$\\
\midrule 

&Baseline & $84.94\pm 2.72$ &${\mathbf{79.57 \pm 3.36}}$&$74.01 \pm 3.99$&$\underline{40.50 \pm 2.87}$&$\underline{87.89 \pm 4.68}$&$\mathbf{69.96 \pm 0.85}$&${73.32 \pm 2.92}$&$\mathbf{54.29 \pm 0.79}$&$\mathbf{45.66 \pm 0.89}$\\
\bottomrule
\end{tabular}
} \vspace{-10px}
\end{table*}

%% file: tables/timetable.tex
\begin{table}[hbt]
\centering
\caption{{Time Costs (in seconds) of Persistent Homology and Kernel-Based Weisfeiler-Lehman Models. VR-based Models Exhibit Significantly Higher Computational Costs Compared to Others.} \label{time: filtration}}
 \resizebox{9cm}{!}
 {
\begin{tabular}{c l c c c c c c c c c}
\toprule
&Model&BZR&COX2&DHFR&ENZYMES&MUTAG&NCI1&PROTEINS&RED-5K&RED-12K\\
\midrule
\parbox[t]{2mm}{\multirow{4}{*}{\rotatebox[origin=c]{90}{TDA}}} &VR-B &$0.69$&$1.16$&$1.87$&$0.69$&$0.06$&$3.61$&$5.16$&$92812.28$&\textcolor{red}{OOR}\\

&VR-L &$1.58$&$2.15$&$3.49$&$2.52$&$0.44$&$12.92$&$8.44$&$93740.33$&$200134.33$\\

&VR-S &$1.36$&$1.96$&$3.15$&$2.67$&$0.36$&$11.07$&$8.84$&$94492.58$&$201988.72$\\
 
&AC-B &$2.17$&$2.75$&$4.56$&$2.93$&$0.44$&$18.43$&$7.19$&$348.87$&$654.76$\\

&AC-L &$3.12$&$3.65$&$6.46$&$3.88$&$0.76$&$26.69$&$9.25$&$393.71$&$750.14$\\

&AC-S &$2.76$&$3.30$&$5.90$&$3.36$&$0.63$&$23.17$&$8.34$&$392.70$&$730.97$\\

\midrule

& WL &$0.14$&$0.18$&$0.51$&$0.33$&$0.04$&$3.38$&$0.59$&$37.77$&$57.94$\\
\midrule
&Baseline &$0.42$&$0.58$&$1.00$&$0.69$&$0.09$&$3.85$&$4.39$&$2539.31$&$5291.97$\\

 \bottomrule

\end{tabular}
}
\end{table}

%% file: relatedwork.tex
We consider three related research areas:  graph kernels, graph neural networks and persistent homology.

\noindent\textbf{Graph Kernels: } The machine learning algorithm that learn by comparing pairs of data points using some particular similarity measures\textemdash \textit{kernels}\textemdash is referred to as the \textit{kernel methods}~(see ~\cite{kriege2020survey} for a recent survey). Typically, a kernel maps the data to a higher dimensional domain where data points of different classes can be better separated. In graphs, a symmetric, positive semidefinite function   is defined on non-empty set of graphs $\mathcal{G}$ to be used as a \textit{graph kernel} with the aim of measuring the similarity between two graphs. 
 
To express kernel function as an inner product in a Hilbert space, let $k$ be a given kernel, there exists a map $\phi: \mathcal{G} \mapsto \mathcal{H}$ into a Hilbert space $\mathcal{H}$ such that $k(G_{1}, G_{2}) = \langle\phi(G_{1}),\phi(G_{2})\rangle$ for all $G_{1}, G_{2} \in \mathcal{G}$ ~\cite{nikolentzos2021graph}. 
As ~\cite{kriege2020survey} outlines, \textquote{early graph kernels such as the shortest path kernel were plagued by worst-case running time complexities that were prohibitively high\textemdash$O(n^4)$\textemdash for large pairs of graphs, with $n$ the largest number of vertices}. 

Over the years, a lot of work have focused on making kernels computationally tractable for large graphs. To this end, ~\cite{shervashidze2011weisfeiler} introduced a kernel method for graphs with discrete labels based on the \textit{Weisfeiler-Leman} (WL) test of isomorphism on graphs. This method employed the \textit{neighbourhood aggregation technique} which functions by assigning an attribute to each vertex based on a summary of its local neighborhood. In an iterative manner, the attributes of the immediate neighbours for each vertex are aggregated to compute a new attribute for the target vertex, which eventually represent the structure of the target's extended neighbourhood. The kernel has a computational complexity of $\mathcal{O}(hm)$ (where $h$ is the number of iterations and $m$ is the maximum number of edges). 

\noindent\textbf{Graph Neural Networks:} While graph kernels provide a good way to work with graph data in a traditional machine learning setting, the attention of machine learning researchers have drifted from hand-crafted features to end-to-end models where both features and the model are learned together ~\cite{cosmo2021graph}. Since graph-based machine learning researchers are focusing on extending deep learning approaches of graph data, the \textit{neural networks} is now a framework of choice to deal with graph data which has led to multiple popular frameworks ~\cite{cosmo2021graph}. ~\cite{sperduti1997supervised} was the first to apply neural networks to structured patterns, which served as a motivation to early studies on \textit{Graph Neural Networks} (GNNs) introduced by ~\cite{gori2005new}. GNNs directly process graphs instead of first representing the graph with a feature vector, as it is done in traditional machine learning methods.   
Inspired by \textit{Convolutional Neural Networks} (CNNs), ~\cite{chen2020convolutional} introduced a generalization of convolutional kernel networks on graph data by proposing a family of multilayer graph kernel representations, which establishes new links between graph CNNs and kernel methods. ~\cite{cosmo2021graph} extended the standard convolution operator to graph domain and designed the Graph Kernel Convolution operation in terms of the inner product between graphs computed through a graph kernel function. For a review on graph neural networks, we refer the reader to a survey by ~\cite{wu2020comprehensive}.

\noindent\textbf{Persistent Homology (PH):} Persistent homology\textemdash an extension of homology\textemdash is regarded as one of the most popular methods in topological data analysis ~\cite{edelsbrunner2000topological, zomorodian2005zomorodian}. PH is capable of bridging gaps between topology and geometry which has provided new opportunities in several fields for researchers. One of the important concepts of PH is to employ a filtration procedure, so that topological generators are equipped with geometric measurements. PH has been used in machine learning since its models are hindered by proper feature representations in their application to high-dimensional complicated systems. PH based machine learning models have been used in different fields which include drug design ~\cite{cang2017topologynet, cang2018integration}, image analysis ~\cite{bae2017beyond}, computational biology ~\cite{pachauri2011topology}, change-point detection ~\cite{islambekov2020harnessing}, to mention a few. For more understanding of PH, we refer interested readers to ~\cite{edelsbrunner2000topological, zomorodian2005zomorodian, zomorodian2008localized}, surveys by ~\cite{pun2018persistent} and ~\cite{edelsbrunner2008persistent}.